\documentclass{article}

\usepackage{arxiv}
\usepackage[utf8]{inputenc} % allow utf-8 input
\usepackage[T1]{fontenc}    % use 8-bit T1 fonts
\usepackage{hyperref}       % hyperlinks
\usepackage{url}            % simple URL typesetting
\usepackage{booktabs}       % professional-quality tables
\usepackage{amsfonts}       % blackboard math symbols
\usepackage{nicefrac}       % compact symbols for 1/2, etc.
\usepackage{microtype}      % microtypography
\usepackage{lipsum}		% Can be removed after putting your text content
\usepackage{graphicx} %figure
\usepackage{natbib}
\usepackage{doi}

\usepackage{colortbl}
\usepackage{tikz}
\usepackage{csquotes} %quotations
\usepackage{amsmath} %equation split
\usepackage{amsfonts} %math font
\usepackage{longtable} %long table
\usepackage[russian,english]{babel}
\usepackage{pdflscape} %landscape
\usepackage{amsmath}

\newcommand{\app}[0]{CSSDetection}

\title{A Survey on Contextualised Semantic Shift Detection \\ 
{\small \textbf{Accepted in ACM Computing Surveys:} {\small \url{https://dl.acm.org/doi/10.1145/3672393}}}}

\author{ \href{https://orcid.org/0000-0002-6594-6644}{\includegraphics[scale=0.06]{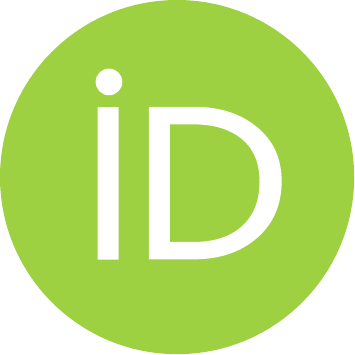}\hspace{1mm}Stefano Montanelli}\thanks{Authors are listed in alphabetical order} \\\\
	Department of Computer Science\\
	Università degli Studi di Milano\\
	Via Celoria 18, 20133 Milan, Italy \\
	\texttt{stefano.montanelli@unimi.it} \\
	\And
	\href{https://orcid.org/0000-0001-8388-2317}{\includegraphics[scale=0.06]{orcid.pdf}\hspace{1mm}Francesco Periti}\thanks{Corresponding author} \\\\ 
	Department of Computer Science\\
	Università degli Studi di Milano\\
		Via Celoria 18, 20133 Milan, Italy \\
	\texttt{francesco.periti@unimi.it} \\
}

\date{}

\renewcommand{\shorttitle}{\textit{arXiv} Template}

\hypersetup{
pdftitle={A Survey on Contextualised Semantic Shift Detection},
pdfsubject={q-cs.CL},
pdfauthor={Stefano Montanelli, Francesco Periti},
pdfkeywords={Computational Semantics, Contextualised Word Embeddings, Semantic Shift Detection},
}

\begin{document}
\maketitle

\begin{abstract}
Semantic Shift Detection (SSD) is the task of identifying, interpreting, and assessing the possible change over time in the meanings of a target word. Traditionally, SSD has been addressed by linguists and social scientists through manual and time-consuming activities. In the recent years, computational approaches based on Natural Language Processing and word embeddings gained increasing attention to automate SSD as much as possible. In particular, over the past three years, significant advancements have been made almost exclusively based on word contextualised embedding models, which can handle the multiple usages/meanings of the words and better capture the related semantic shifts. In this paper, we survey the approaches based on contextualised embeddings for SSD (i.e., \app) and we propose a classification framework characterised by {\em meaning representation}, {\em time-awareness}, and {\em learning modality} dimensions. The framework is exploited i) to review the measures for shift assessment, ii) to compare the approaches on performance, and iii) to discuss the current issues in terms of scalability, interpretability, and robustness. Open challenges and future research directions about \app\ are finally outlined.
\end{abstract}


\section{Introduction}
Word meanings in a language are influenced over time by social practices, events, and political circumstances~\citep{keidar2022slangvolution,periti2022semantic,azarbonyad2017words}. Detecting, interpreting, and assessing the possible change of a word over time are usually considered as steps of a broader {\em Semantic Shift Detection} (SSD) task. Capturing semantic shift requires to arrange testing procedures as well as to define and standardise interviews that are eventually exploited to build large catalogues of word descriptions. All this work is generally addressed by interested scholars, like linguists and social scientists, through manual and time-consuming approaches of ``close reading'' that keep humans ``in-the-loop''.

The growing attention on Computational Semantics issues as well as the recent availability of large digitised diachronic corpora in many different languages, like English~\citep{alatrash2020ccoha}, Swedish~\citep{adesam2019exploring}, German~\citep{schlechtweg2020semeval}, Latin~\citep{mcgillivray2013tools}, Italian~\citep{basile2019kronositad}, Russian~\citep{kutuzov2021three},
Spanish~\citep{zamora2022lscdiscovery},
Chinese~\citep{chen2022lexicon}, and Norwegian~\citep{kutuzov2022nordiachange}, pushed the emergence of a novel family of approaches based on Natural Language Processing (NLP) techniques to automate the SSD task as much as possible.

In this context, distributional word representations (i.e., word embeddings) emerged as an effective solution, based on the idea that semantically related words are close to each other in the embedding space~\citep{mikolov2013efficient}. The use of static embedding solutions is widely adopted and the main approaches have been reviewed in three survey papers~\citep{tang2018state,kutuzov2018diachronic,tahmasebi2018survey}. Typically, static embeddings are used to detect how a word changes in its dominant sense, without considering the possible additional, subordinate senses/meanings that the word can have. However, subordinate senses can change on their own regardless of their dominant sense. For example, considering the word \verb|rock|, the \verb|music| meaning evolved over time to encompass both \verb|music| and a particular lifestyle, while the \verb|stone| meaning remained unchanged~\citep{tahmasebi2013models,mitra2015automatic}. This issue has motivated the recent efforts to enforce SSD by leveraging contextualised embeddings, which are capable of handling the so-called \textit{colexification} phenomena such as homonymy and polysemy. As a result, SSD solutions based on contextualised embeddings have emerged, but a classification framework and a corresponding survey of existing approaches are still missing.

In this paper, we define \app\ as a {\em Contextualised Semantic Shift Detection} task and we survey the main approaches to SSD addressed through the use of contextualised embedding techniques. To this end, we propose a classification framework based on three dimensions of analysis, namely {\em meaning representation}, {\em time-awareness}, and {\em learning modality}, that allows to effectively describe the featuring properties of both \textit{form}- and \textit{sense}-oriented approaches in which \app\ solutions are typically distinguished. Assessment methods and metrics used for \app\ are also surveyed to discuss how the detected semantic shift of a word is measured and quantified by the considered approaches.

In the recent years, a growing attention has been captured by SSD issues, and a number of events with competitive shared tasks and corpora have been proposed, such as  SemEval-20 Task 1~\citep{schlechtweg2020semeval}, DIACRIta-20~\citep{basile2020diacr}, RuShiftEval-21~\citep{kutuzov2021rushifteval}, and LSCDiscovery-22~\citep{zamora2022lscdiscovery}. As a further contribution of our survey, the \app\ approaches are compared according to their results in these competitions (where available) with the aim to discuss the related performance issues and possible limitations in real applications. 

Unlike the existing surveys on static SSD, the goal of our survey is to focus on contextualised approaches and to highlight the computational perspective, rather than the linguistics one.

The paper is organised as follows. Section~\ref{sec:problem} presents the \app\ problem and the related workflow with related formalisation. The proposed survey framework for approach classification is illustrated in Section~\ref{sec:classification}. The classification of state-of-the-art approaches is discussed in Section~\ref{sec:approaches}. A comparative analysis of approach performance is provided in~\ref{sec:comparison}; issues about scalability, interpretability, and robustness of \app\ approaches are discussed in Section~\ref{sec:limitations}. Finally, in Section~\ref{sec:conclusion}, we outline the open challenges and we give our concluding remarks.

\section{Problem statement}\label{sec:problem}
Consider a diachronic document corpus $\mathcal{C} = \bigcup^{i=n}_{i=1} C_i$ where $C_i$ denotes a set of documents (e.g., sentences, paragraphs) of the time $t_i$. \app\ consists in assessing the change of meaning for a set of target words $\mathcal{W}$ occurring in $\mathcal{C}$ across the whole time span $[1 \dots n]$ by leveraging contextualised embeddings. 

Approaches to \app\ rely on semantic modeling of words and their foundations lie in the well-known distributional hypothesis: \blockquote{You shall know a word by the company it keeps}~\citep{firth1957synopsis}, meaning that the semantic representation of a word is determined by analysing the patterns of lexical co-occurrence within a considered document corpus.

As a difference with static models, where words are encoded in a single vector representation, contextualised models generate different word representations according to the context in which they occur. For instance, consider the word \verb|sex|. Different semantic vectors are generated when the word in the input sequence is used with the \verb|fair sex| connotation or with the \verb|sexual activity| meaning.

For the sake of readability, in the following, we consider the problem of \app\ on a set of documents $\mathcal{C} = C_1 \cup C_2$ and we consider to evaluate the shift of a given target word $w \in \mathcal{W}$ on a single time period from $t_1$ to $t_2$. This simplification enables to review approaches to \app\ in a clear and concise fashion, while being easily extendable to the general case. As a matter of fact, when $\mathcal{C}$ spans more than two time periods, the change is typically measured by re-implementing the approaches across all contiguous pairs of periods~\citep{giulianelli2020analysing}.

\begin{figure}[!ht]
    \centering
    \includegraphics[width=0.5\columnwidth]{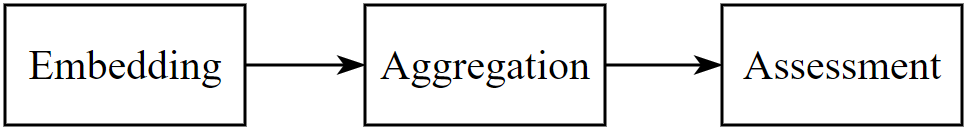}
    \caption{A general workflow of \app}
    \label{fig:workflow}
\end{figure}

The \app\ workflow is described in Figure~\ref{fig:workflow}.

The initial {\em embedding stage} has the goal to represent the word occurrences in a multi-dimensional semantic space where the word representations are similar for word occurrences in similar sentences. An optional {\em aggregation stage} can be enforced to group multiple word representations into a single one for detecting similar usage and/or reducing the computational complexity of the overall \app\ task. 
For example, word occurrences can be aggregated according to a sense-oriented criterion. As another example, multiple word representations can be synthesised into a single prototype representation. The final {\em assessment stage} consists in the application of a semantic measure to evaluate how the meanings of the word shifted over time.\\

\paragraph{Embedding.} 
Consider the subsets of documents $C_w^1 \subseteq C_1$ and $C_w^2 \subseteq C_2$ that contain the word $w$. In the embedding stage, a contextualised model $m$ (e.g.,~BERT~\citep{devlin2019bert}, RoBERTA~\citep{yinhan2019roberta}, ELMo~\citep{peters2018deep}) is employed to extract an embedding vector for each occurrence of $w$ in $C_w^1$ and $C_w^2$. The contextualised embedded representation of the word $w$ in the $i$-th document of a corpus $C_w^j$ is denoted by $e_{w,i}^{j}$ ($j \in \{1, 2\}$). Then, the representation of the word $w$ in a corpus $C_w^j$ is defined as:
$\Phi_w^j = \{ e_{w,1}^{j}, \dots, e_{w,z}^{j} \}$, with $z$ being the cardinality of $C_w^j$, namely the number of documents in $C_w^j$.
As a result, we denote as $\Phi_w^1$ and $\Phi_w^2$ the sets of embedding vectors generated for the word $w$ at time $t_1$ and $t_2$, respectively.\\

\paragraph{Aggregation.} 
This stage is optionally executed and it has two main goals: i) to recognise when different word occurrences purport a similar meaning, and ii) to reduce the number of elements to consider for shift detection.
To this end, clustering and averaging techniques are proposed for aggregating the word embeddings previously created.\\

\begin{figure}[!ht]
    \centering
    \includegraphics[width=0.5\columnwidth]{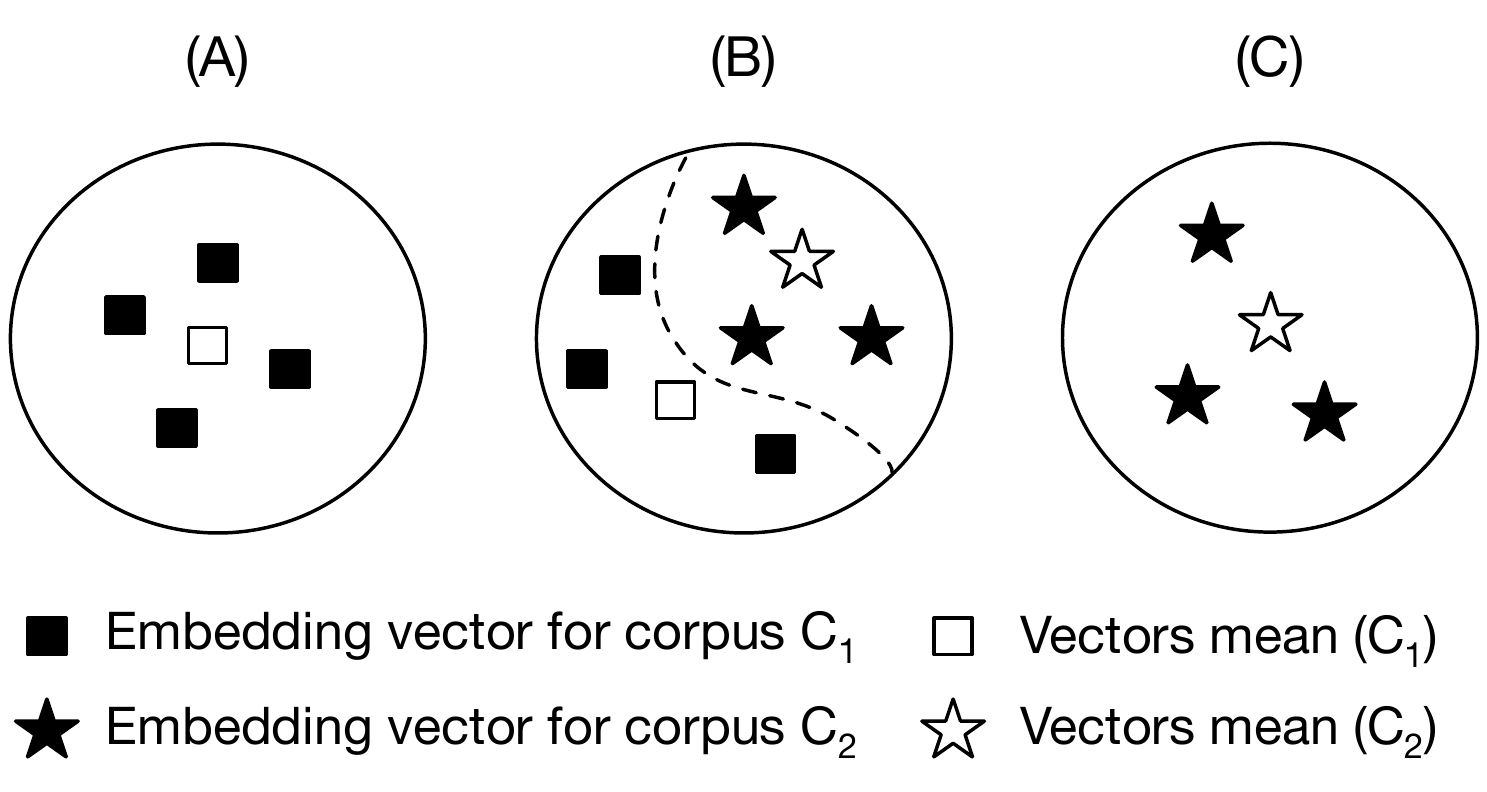}
    \caption{Possible cluster composition (from~\cite{periti2022what})}
    \label{fig:clusters}
\end{figure}

\textit{Clustering.} \ \ Clustering techniques are employed to group similar word embeddings in a cluster, each one loosely denoting a specific word meaning. In some approaches, it is assumed that the corpus is {\em static}, meaning that all the documents in $C_w^1$ and $C_w^2$ are available as a whole. Then, a \textit{joint} clustering operation is executed over the embeddings of $\Phi_w^1 \cup \Phi_w^2$ (e.g.~\cite{martinc2020capturing}). In other approaches, it is assumed that the corpus is {\em dynamic}, meaning that documents become available at different time moments and a \textit{separate} clustering operation is performed over the embeddings of $\Phi_w^1$ and $\Phi_w^2$, individually (i.e., one exclusively on $\Phi_w^1$ and another exclusively on $\Phi_w^2$ embeddings). When a separate clustering is executed, the resulting clusters need to be aligned in order to recognise similar word meanings at different consecutive times (e.g.~\cite{kanjirangat2020sst}). To overcome the need for aligning clusters, an \textit{incremental} clustering operation is employed to progressively group the embedding available at the different time steps (e.g.~\cite{periti2022what}). The result of clustering is a set of $k$ clusters where the $i$-th cluster is denoted as $\phi_{w,i}$ and it can fall into one of the following cases (see Figure~\ref{fig:clusters}): 
\begin{enumerate}
    \item [A.]$\phi_{w,i}$ contains only embeddings from $C_w^1$;
    \item [B.]$\phi_{w,i}$ contains a mixture of embeddings from both $C_w^1$ and $C_w^2$;
    \item [C.]$\phi_{w,i}$ contains only embeddings from $C_w^2$.
\end{enumerate}
As a result, a cluster $\phi_{w,i} = \phi_{w,i}^1 \cup \phi_{w,i}^2$ is composed by the union of two partitions $\phi_{w,i}^1$ and $\phi_{w,i}^2$ denoting the embeddings from $\Phi_w^1$ and $\Phi_w^2$, respectively. When a \textit{joint} or \textit{incremental} clustering is applied, the resulting clusters can belong to any of the above cases (i.e., A, B, and C). When a \textit{separate} clustering is applied, the resulting clusters can just belong to A and C cases, meaning that $\phi_{w,i}^2 = \emptyset$ and $\phi_{w,i}^1 = \emptyset$, respectively.\\

\textit{Averaging.} \ \ Averaging techniques consist in determining a prototypical representation of the word $w$. As an option, a \textit{word}-prototype, can be computed by averaging all its embedding. In this case, \textit{word}-prototypes $\mu_w^1$ and $\mu_w^2$ are created as the average embeddings of $\Phi_w^1$ and $\Phi_w^2$, respectively~(e.g.,~\cite{rodina2020elmo}). As an alternative option, averaging can be executed on top of the results of clustering. For each cluster, averaging is used to create a prototypical representation of all the cluster elements (i.e., the centroid of the cluster). In particular, \textit{sense}-prototypes $c_{w,i}^1$, $c_{w,i}^2$ can be created for each cluster $\phi_{w,i}$ as the average embedding of its cluster partitions $\phi_{w,i}^1, \phi_{w,i}^2$, respectively~(e.g.,~\cite{periti2022what}).\\ 

\textbf{Assessment.} 
This stage has the goal to measure the shift on the meanings of the word $w$ across the corpora $C_1$ and $C_2$ by considering the sets $\Phi_w^1$ and $\Phi_w^2$. In the literature, a number of functions are proposed for semantic shift assessment. We can distinguish measures that assess the shift by considering the whole set of embedding representations $\Phi_w^i$, by those that exploit the prototypical representations $c_w^i$ and/or $\mu_w^i$ generated during the aggregation step through clustering and/or averaging.  
According to~\cite{kutuzov2018diachronic}, the definition of a rigorous, formal, mathematical model for representing the assessment functions used in \app\ approaches is a challenging issue. In the following, we provide a formal definition of an abstract function $f$ that has the goal to encompass all the existing assessment measures. The semantic shift assessment $s_w=f(\cdot,\cdot)$ is defined as follows: 
\begin{equation*}
    f:\{\mathbb{R}^D\}^{(p_1 + z_1\cdot\delta)}, \{\mathbb{R}^D\}^{(p_2 + z_2\cdot\delta)}, \ c \rightarrow \mathcal{S}
\end{equation*}
where $D$ is the dimension of the word vectors in $\Phi_w^1$ and $\Phi_w^2$; $p_1, p_2$ are the number of prototypical embeddings under consideration for $C_w^1, C_w^2$, respectively; $z_1, z_2$ are the number of vectors in $\Phi_w^1$ and $\Phi_w^2$, respectively; $\delta \in \{0, 1\}$ is a flag that allows to distinguish the approaches according to the kind of embedding used (i.e., original and/or prototypical); $c$ is a counting function that determines the normalised number of embeddings in the cluster partitions $\phi_{w,i}^1$ and $\phi_{w,i}^2$, respectively.

The counting function $c$ is defined as:
\begin{equation*}
    c:\{\mathbb{R}^D\}^{z_1}, \{\mathbb{R}^D\}^{z_2} \rightarrow \mathbb{R}^k,\mathbb{R}^k
\end{equation*}
where $k$ denotes the number of $k$ clusters obtained when a clustering operation is enforced during the aggregation stage.  When the clustering operation is not enforced, each embedding is mapped to a singleton group (i.e., $k=z_1 + z_2$).

The signature of $f$ depends on the possible execution of an aggregation technique:
\begin{itemize}
    \item {\em Clustering}. When the clustering operation is executed, then $p_1=p_2=0$ and $\delta = 1$. This means that all the $z_1 + z_2$ embeddings in $\Phi_w^1 \cup \Phi_w^2$ are exploited for semantic shift assessment~(e.g.,~\cite{martinc2020capturing}).
     
    \item {\em Averaging}. When the averaging operation is executed, then $p_1 = p_2 = 1$. In some approaches, $\delta=0$ and this means that the function $f$ is defined as a distance measure over prototypical representations~(e.g.,~\cite{martinc2020leveraging}). In some other approaches, $\delta=1$ and this means that $f$ is defined as a distance measure over the original embeddings $\Phi_w$ and their prototypical representations~(e.g.,~\cite{pomsl2020circe}).
    
    \item {\em Clustering + Averaging}. When both clustering and averaging are performed, $p_1, p_2 > 0$ and $\delta$ can be both $0$ or $1$ as in the previous case~(e.g.,~\cite{periti2022what}).

\end{itemize}

The output $\mathcal{S}$ is defined according to four different assessment questions.
\begin{itemize}

\item{\em Grade Change Detection.}
The goal of Grade Change Detection is to quantify the assessment $s_w$. Then, $\mathcal{S}=\mathbb{R}$ represents the extent to which $w$ shifts between $C_1$ and $C_2$~\citep{schlechtweg2020semeval}. 

\item{\em Binary Change Detection.}
The goal of Binary Change Detection is to classify $w$ as stable (without lost or gained sense(s)) or changed (with lost or gained sense(s)). In this case, $s_w$ is binary, meaning that $\mathcal{S} = \{0,1\}$ for stable and changed, respectively~\citep{schlechtweg2020semeval}.

\item{\em Sense Gain Detection.}
The goal of Sense Gain Detection is to recognise whether $w$ gained meanings or not. In this case, $s_w$ is binary, meaning that $\mathcal{S} = \{0,1\}$ for not-gained and gained, respectively~\citep{zamora2022lscdiscovery}.

\item{\em Sense Loss Detection.}
The goal of Sense Gain Detection is to recognise whether $w$ lost meanings or not. In this case, $s_w$ is binary, meaning that $\mathcal{S} = \{0,1\}$ for not-lost and lost, respectively~\citep{zamora2022lscdiscovery}.
\end{itemize}

In this survey, we focus on approaches that adopt Grade Change Detection since it is the most commonly enforced assessment. As a matter of fact, we note that the approaches based on Grade Change Detection can be transformed into Binary Change Detection by binarising $s_w$ through a threshold $\theta$. We do not address Sense Gain and Sense Loss Detection as they are relatively novel assessment questions. 

%Furthermore, it is worth noting that other problems related to the semantic shifts assessment are being investigated through the use of contextualized embeddings. For example, detecting borrowings and cognates~\citep{fourrier2022caveats}, tracing the evolution of word meanings~\citep{periti2022what}, and sentence time prediction~\citep{rosin2021time}.

For the sake of clarity, we summarise the notation used throughout this paper in Table~\ref{tab:notation}.
\begin{table}[!h]
\centering
\resizebox{0.65\textwidth}{!}{%
\begin{tabular}{cl}
\textbf{Notation}                   & \textbf{Definition}                                                             \\ \hline
%\multicolumn{1}{|c|}{$\mathcal{C}$} & \multicolumn{1}{l|}{Diachronic document corpus}                                 \\ \hline
%\multicolumn{1}{|c|}{$t_j$}         & \multicolumn{1}{l|}{Time $j$-th}                                                \\ \hline
\multicolumn{1}{|c|}{$w$}           & \multicolumn{1}{l|}{Target word}                                                \\ \hline
\multicolumn{1}{|c|}{$C_j$}         & \multicolumn{1}{l|}{Set of documents at time $t_j$}                         \\ \hline
%\multicolumn{1}{|c|}{$\mathcal{W}$} & \multicolumn{1}{l|}{Set of target words}                                        \\ \hline
\multicolumn{1}{|c|}{$C_w^j$}       & \multicolumn{1}{l|}{Subset of documents of $C_j$ containing the word $w$}  \\ \hline
\multicolumn{1}{|c|}{$e_{w,i}^{j}$}  & \multicolumn{1}{l|}{Contextualised embedding of the word $w$ in the $i$-th document of a corpus $C_w^j$} \\ \hline
\multicolumn{1}{|c|}{$\Phi_w^j$}    & \multicolumn{1}{l|}{Set of the embeddings of $w$ in the corpus $C_w^j$}          \\ \hline
\multicolumn{1}{|c|}{$\phi_{w,i}$}  & \multicolumn{1}{l|}{$i$-th cluster containing the embeddings of the word $w$}  \\ \hline
\multicolumn{1}{|c|}{$\phi_{w,i}^j$} & \multicolumn{1}{l|}{Subset of contextualised embeddings $\Phi_w^j$ in the cluster $\phi_{w,i}$}            \\ \hline
\multicolumn{1}{|c|}{$\mu_w^j$}     & \multicolumn{1}{l|}{Prototypical representation of $w$ for $\Phi_w^j$} \\ \hline
\multicolumn{1}{|c|}{$c_{w,i}^j$}   & \multicolumn{1}{l|}{Prototypical representation of $w$ for $\phi_{w,i}^j$}      \\ \hline
\end{tabular}%
}
\caption{A reference table of notations used in the paper}
\label{tab:notation}
\end{table}

\section{A classification framework for \app}\label{sec:classification}
A consolidated and widely-accepted classification framework of \app\ approaches is not available. A basic framework is focused on the meaning representation of the words by distinguishing between form- and sense-based approaches~\citep{giulianelli2020analysing,qiu2022histbert}. However, such a distinction is not universally recognised with a unique interpretation. Sometimes, these two categories are %used to distinguish approaches where static embeddings are employed (form-based) from approaches featured by the use of contextualised embeddings (sense-based)~\citep{tahmasebi2018survey}. Form- and sense-based categories are also
referred as \textit{type-} and \textit{token}-based, where averaging and clustering are enforced to aggregate embeddings,  respectively~\citep{laicher2020clims,schlechtweg2020semeval}. More recently, \textit{average}- and \textit{cluster}-based categories have been proposed to rename form and sense ones to highlight the method used for embedding aggregation~\citep{periti2022what}. \\

In the following, we propose a comprehensive classification framework that extends the basic distinction between form- and sense-based approaches by introducing three dimensions of analysis, namely {\em meaning representation}, {\em time-awareness}, and {\em learning modality} (see Table~\ref{tab:cls}).  
\begin{table}[h]
\centering
\resizebox{0.5\textwidth}{!}{%
\begin{tabular}{|c|c|c|}
\hline
\textbf{Meaning representation} & \textbf{Time-awareness} & \textbf{Learning modality} \\ \hline
form-based       & time-oblivious                   & supervised      \\

sense-based     & time-aware                    & unsupervised \\ \hline
\end{tabular}%
}
\caption{A classification framework for \app}
\label{tab:cls}
\end{table}\\

\textbf{Meaning representation.}
Borrowing the distinction proposed by~\cite{giulianelli2020analysing}, this dimension focuses on the meaning representation of a word. Two categories are defined:
\begin{itemize}

    \item \textit{form-based}: the meaning representation concerns the high-level properties of the target word $w$, such as its degree of polysemy or its dominant sense. When the polysemy is considered, the \app\ approaches do not enforce any aggregation stage and the semantic shift of $w$ is assessed by measuring the degree of change on the embeddings $\Phi_w^1$ and $\Phi_w^2$ (i.e., change on the degree of polysemy). When the dominant sense is considered, all the meanings of $w$ are collapsed into a single one on which the shift is assessed. Typically, the embeddings $\Phi_w^1$ and $\Phi_w^2$ are averaged into corresponding word prototypes $\mu_w^1$ and $\mu_w^2$, respectively. In this case, the \app\ approaches focus on one meaning of $w$ that can be considered as an approximation of the {\em dominant sense} since, generally, it is the most frequent in the corpus, and thus the one most represented in the word prototype. We stress that form-based approaches are not able to represent how minor meanings \textit{compete} and \textit{cooperate} to change the dominant sense~\citep{hu2019diachronic}.
 
    \item \textit{sense-based}: the meaning representation concerns the low-level properties of the target word $w$, such as its different word usages (i.e., its multiple meanings). All the senses of a word $w$ are represented and considered in the shift assessment, namely the dominant sense and the minor ones. Typically, the embeddings $\Phi_w^1$ and $\Phi_w^2$ are aggregated into clusters, each one representing a different usage/meaning of $w$. Sense-based approaches allow to capture the changes over the different meanings of $w$ as well as to interpret the word change (e.g., a new/existing meaning has gained/lost importance).
\end{itemize}

\textbf{Time awareness.}
This dimension focuses on how the time information of the documents is considered in the embedding model. Two categories are defined:
\begin{itemize}
    \item \textit{time-oblivious}: this category is based on the assumption that a document of time $t$ adopts linguistic patterns that already characterise the language at the time $t$ by its own. Thus, it is not needed that the embedding model is aware of the time in which a document is inserted in the corpus. A time-oblivious model is based on \textit{the contextual nature of embeddings generated by the model, which by definition are dependent on the context that is always time-specific}~\citep{martinc2020capturing}. 
    
    \item \textit{time-aware}: this category is based on the assumption that contextualised embedding models are not capable of \textit{adapting to time and generalising temporally} since they are \textit{usually pre-trained on corpora derived from a snapshot of the web crawled at a specific moment in time}~\citep{rosin2021time}. Thus, it is needed that the embedding model is aware of the time in which a document is inserted in the corpus. As a result, a time-aware model encodes the time information as well as the linguistic context of a document while generating the embeddings.
\end{itemize}

\textbf{Learning modality.}
This dimension is about the possible use of external knowledge for describing and learning the word meanings to recognise. Two categories are defined:
\begin{itemize}
    \item \textit{supervised}: a form of supervision is enforced to inject external knowledge to support the shift assessment. In addition to the text in the corpora $C_1$ and $C_2$, a lexicographic/manual supervision is employed. By lexicographic supervision, we mean that a dictionary/thesaurus is introduced to recognise the meanings of the word $w$. This solution can be considered as an alternative to aggregation by clustering for meaning identification. By manual supervision, we mean that a human-annotated dataset with gold labels is provided for training the embedding model. 
    
    \item \textit{unsupervised}: the shift assessment is exclusively based on the text of the corpora $C_1$, $C_2$ without any external knowledge support. As a result, the word meanings to recognise emerge from the corpora and the shift is completely assessed by exploiting unsupervised learning techniques. The use of aggregation by clustering is an example of unsupervised learning for meaning detection.
\end{itemize} 

\section{Approaches to \app}\label{sec:approaches}
In this section, we review the literature about \app\ according to the classification framework discussed in Section~\ref{sec:classification}. In particular, the solutions are presented in Sections~\ref{sec:form} and~\ref{sec:sense} according to the meaning representation of the considered target word, namely \textit{form-} and \textit{sense-} based approaches, respectively. Moreover, in Section~\ref{sec:ensemble}, we describe the so-called \textit{ensemble} approaches, namely approaches that are based on a combination of form-/sense-based solutions. 

For the sake of comparison, in each category (i.e., form, sense, ensemble), a summary table is provided to frame the literature papers according to our dimensions of analysis as well as to report additional descriptive features about the following aspects:   
\begin{itemize}
    \item \textit{Language model}: the contextualise language model used (e.g., ELMo, BERT, RoBERTa);
    \item \textit{Training language}: the language of the dataset used for training the model. The possible options are {\em monolingual} to denote when training is executed on a single language, or {\em multilingual} when more than one language is considered.
    \item \textit{Type of training}: how the model is trained. We distinguish five categories: 
    \begin{itemize}
        \item \textit{trained}: the model is trained from scratch through the typical objective function of the architecture model;
        
        \item \textit{pre-trained}: the model is pre-trained through the typical objective function without further training;
        
        \item fine-tuned for \textit{domain-adaptation}: the model is pre-trained through the typical objective function, then it is fine-tuned on new data through the same objective function;
        
        \item fine-tuned for \textit{incremental domain-adaptation}:  the model is fine-tuned on the corpus of the first time period $C_1$.
        Then, it is re-tuned separately on the corpus $C_2$. The model at time $t_2$ is initialised with the weights from the model at time $t_1$, so that both models are inherently related the one to the other; %Then, it is \textit{re}-tuned separately on each other corpus $C_{i}$ with $i \in \{2, \dots n\}$. Each model at time $t_{i}$ is initialised with the weights from the model at time $t_{i-1}$, so that all the models are inherently related the one to the others; 
        \item \textit{fine-tuned}: the model is pre-trained through the typical objective function, then it is fine-tuned on new data through a different objective function.
    \end{itemize}
    
    \item \textit{Layer}: the architecture's layer(s) from which word representations are extracted;
    
    \item \textit{Layer aggregation}: the type of aggregation used to synthesise the word representations extracted from different layers into a single embedding;
    
    \item \textit{Clustering algorithm}: the clustering algorithm used in the aggregation stage;
    
    \item \textit{Shift function}: the function $f$ used to detect/assess the semantic shift;
    
    \item \textit{Corpus language}: the natural language of the corpus in the considered experiments of shift assessment (e.g., English, Italian, Spanish).
\end{itemize}

\subsection{Form-based approaches}\label{sec:form}

\begin{table}[!ht]
\centering
\resizebox{\textwidth}{!}{%
\begin{tabular}{ccccccccccc}
\textbf{Ref.} &
  \textbf{\begin{tabular}[c]{@{}c@{}}Time\\ awareness\end{tabular}} &
  \textbf{\begin{tabular}[c]{@{}c@{}}Learning\\ modality\end{tabular}} &
  \textbf{\begin{tabular}[c]{@{}c@{}}Language\\ model\end{tabular}} &
  \textbf{\begin{tabular}[c]{@{}c@{}}Training\\ language\end{tabular}} &
  \textbf{\begin{tabular}[c]{@{}c@{}}Type of \\ training\end{tabular}} &
  \textbf{Layer} &
  \textbf{\begin{tabular}[c]{@{}c@{}}Layer \\ aggregation\end{tabular}} &
  \textbf{\begin{tabular}[c]{@{}c@{}}Clustering\\ algorithm\end{tabular}} &
  \textbf{\begin{tabular}[c]{@{}c@{}}Shift \\ function\end{tabular}} &
  \textbf{\begin{tabular}[c]{@{}c@{}}Corpus \\ language\end{tabular}} \\ \hline
\multicolumn{1}{|c}{\citeauthor{arefyev2021deep}} &
  time-oblivious &
  supervised &
  XLM-R-large &
  multilingual &
  fine-tuned &
  last &
  - &
  - &
  APD &
  \multicolumn{1}{c|}{Russian} \\ \hline
\multicolumn{1}{|c}{\citeauthor{beck2020diasense}} &
  time-oblivious &
  unsupervised &
  mBERT-base &
  multilingual &
  pre-trained &
  last two &
  average &
  K-Means &
  CD &
  \multicolumn{1}{c|}{\begin{tabular}[c]{@{}c@{}}English,\\ German,\\ Latin,\\ Swedish\end{tabular}} \\ \hline
\multicolumn{1}{|c}{\citeauthor{martinc2020leveraging}} &
  time-oblivious &
  unsupervised &
  \begin{tabular}[c]{@{}c@{}}BERT-base,\\ mBERT-base\end{tabular} &
  \begin{tabular}[c]{@{}c@{}}monolingual,\\ multilingual\end{tabular} &
  domain-adaptation &
  last four &
  sum &
  - &
  CD &
  \multicolumn{1}{c|}{\begin{tabular}[c]{@{}c@{}}English,\\ Slovenian\end{tabular}} \\ \hline
\multicolumn{1}{|c}{\citeauthor{horn2021exploring}} &
  time-oblivious &
  unsupervised &
  \begin{tabular}[c]{@{}c@{}}BERT-base,\\ RoBERTa-base\end{tabular} &
  monolingual &
  \begin{tabular}[c]{@{}c@{}}domain-adaptation,\\ pre-trained\end{tabular} &
  - &
  - &
  - &
  CD &
  \multicolumn{1}{c|}{English} \\ \hline
\multicolumn{1}{|c}{\citeauthor{hofmann2021dynamic}} &
  time-aware &
  unsupervised &
  BERT-base &
  monolingual &
  fine-tuned &
  last &
  - &
  - &
  CD &
  \multicolumn{1}{c|}{English} \\ \hline
\multicolumn{1}{|c}{\citeauthor{zhou2020temporalteller}} &
  time-aware &
  unsupervised &
  BERT-base &
  monolingual &
  domain-adaptation &
  last four &
  sum &
  - &
  CD &
  \multicolumn{1}{c|}{\begin{tabular}[c]{@{}c@{}}English,\\ German,\\ Latin,\\ Swedish\end{tabular}} \\ \hline
\multicolumn{1}{|c}{\citeauthor{rosin2021time}} &
  time-aware &
  unsupervised &
  \begin{tabular}[c]{@{}c@{}}BERT-base,\\ BERT-tiny\end{tabular} &
  monolingual &
  fine-tuned &
  \begin{tabular}[c]{@{}c@{}}all,\\ last,\\ last four\end{tabular} &
  average &
  - &
  \begin{tabular}[c]{@{}c@{}}CD, \\ TD\end{tabular} &
  \multicolumn{1}{c|}{\begin{tabular}[c]{@{}c@{}}English,\\ Latin\end{tabular}} \\ \hline
\multicolumn{1}{|c}{\citeauthor{rosin2022temporal}} &
  time-aware &
  unsupervised &
  \begin{tabular}[c]{@{}c@{}}BERT-base,\\ BERT-small,\\ BERT-tiny\end{tabular} &
  monolingual &
  fine-tuned &
  \begin{tabular}[c]{@{}c@{}}all,\\ last, \\ last four, \\ last two\end{tabular} &
  average &
  - &
  CD &
  \multicolumn{1}{c|}{\begin{tabular}[c]{@{}c@{}}English,\\ German,\\ Latin\end{tabular}} \\ \hline
\multicolumn{1}{|c}{\citeauthor{kutuzov2020uio}} &
  time-oblivious &
  unsupervised &
  \begin{tabular}[c]{@{}c@{}}BERT-base,\\ ELMo,\\ mBERT-base\end{tabular} &
  \begin{tabular}[c]{@{}c@{}}monolingual,\\ multilingual\end{tabular} &
  \begin{tabular}[c]{@{}c@{}}domain-adaptation,\\ incremental domain-adaptation,\\ pre-trained,\\ trained\end{tabular} &
  \begin{tabular}[c]{@{}c@{}}all,\\ last,\\ last four\end{tabular} &
  average &
  - &
  \begin{tabular}[c]{@{}c@{}}APD,\\ CD, \\ PRT\end{tabular} &
  \multicolumn{1}{c|}{\begin{tabular}[c]{@{}c@{}}English,\\ German,\\ Latin,\\ Swedish\end{tabular}} \\ \hline
\multicolumn{1}{|c}{\citeauthor{giulianelli2020analysing}} &
  time-oblivious &
  unsupervised &
  BERT-base &
  monolingual &
  pre-trained &
  all &
  sum &
  - &
  APD &
  \multicolumn{1}{c|}{English} \\ \hline
\multicolumn{1}{|c}{\citeauthor{keidar2022slangvolution}} &
  time-oblivious &
  unsupervised &
  RoBERTa-base &
  monolingual &
  domain-adaptation &
  \begin{tabular}[c]{@{}c@{}}all,\\ first, \\ last\end{tabular} &
  sum &
  - &
  APD &
  \multicolumn{1}{c|}{English} \\ \hline
\multicolumn{1}{|c}{\citeauthor{pomsl2020circe}} &
  time-aware &
  unsupervised &
  \begin{tabular}[c]{@{}c@{}}BERT-base,\\ mBERT-base\end{tabular} &
  \begin{tabular}[c]{@{}c@{}}monolingual,\\ multilingual\end{tabular} &
  fine-tuned &
  last &
  - &
  - &
  APD &
  \multicolumn{1}{c|}{\begin{tabular}[c]{@{}c@{}}English,\\ German,\\ Latin,\\ Swedish\end{tabular}} \\ \hline
\multicolumn{1}{|c}{\citeauthor{kudisov2022bos}} &
  time-oblivious &
  unsupervised &
  XLM-R-large &
  multilingual &
  pre-trained &
  - &
  - &
  - &
  APD &
  \multicolumn{1}{c|}{Spanish} \\ \hline
\multicolumn{1}{|c}{\citeauthor{laicher2021explaining}} &
  time-oblivious &
  unsupervised &
  BERT-base &
  monolingual &
  pre-trained &
  \begin{tabular}[c]{@{}c@{}}first, \\ first + last,\\ first four,\\ last,\\ last four\end{tabular} &
  average &
  - &
  \begin{tabular}[c]{@{}c@{}}APD,\\ APD-OLD/NEW,\\ CD\end{tabular} &
  \multicolumn{1}{c|}{\begin{tabular}[c]{@{}c@{}}English,\\ German,\\ Swedish\end{tabular}} \\ \hline
\multicolumn{1}{|c}{\citeauthor{wang2020university}} &
  time-oblivious &
  unsupervised &
  mBERT-base &
  multilingual &
  pre-trained &
  last &
  - &
  - &
  \begin{tabular}[c]{@{}c@{}}APD,\\ HD\end{tabular} &
  \multicolumn{1}{c|}{Italian} \\ \hline
\multicolumn{1}{|c}{\citeauthor{kutuzov2020distributional}} &
  time-oblivious &
  unsupervised &
  \begin{tabular}[c]{@{}c@{}}BERT-base,\\ BERT-large,\\ ELMo,\\ mBERT-base\end{tabular} &
  \begin{tabular}[c]{@{}c@{}}monolingual,\\ multilingual\end{tabular} &
  \begin{tabular}[c]{@{}c@{}}domain-adaptation,\\ pre-trained\end{tabular} &
  \begin{tabular}[c]{@{}c@{}}all, \\ last, \\ last four\end{tabular} &
  average &
  - &
  \begin{tabular}[c]{@{}c@{}}APD,\\ DIV,\\ PRT\end{tabular} &
  \multicolumn{1}{c|}{\begin{tabular}[c]{@{}c@{}}English,\\ German,\\ Latin,\\ Swedish,\\ Russian\end{tabular}} \\ \hline
\multicolumn{1}{|c}{\citeauthor{ryzhova2021detection}} &
  time-oblivious &
  unsupervised &
  \begin{tabular}[c]{@{}c@{}}ELMo,\\ RuBERT~\citeyear{kuratov2019adaptation}\end{tabular} &
  multilingual &
  \begin{tabular}[c]{@{}c@{}}pre-trained,\\ trained\end{tabular} &
  - &
  - &
  - &
  APD &
  \multicolumn{1}{c|}{Russian} \\ \hline
\multicolumn{1}{|c}{\citeauthor{rodina2020elmo}} &
  time-oblivious &
  unsupervised &
  \begin{tabular}[c]{@{}c@{}}ELMo,\\ RuBERT\end{tabular} &
  \begin{tabular}[c]{@{}c@{}}monolingual,\\ multilingual\end{tabular} &
  domain-adaptation &
  last &
  - &
  - &
  PRT &
  \multicolumn{1}{c|}{Russian} \\ \hline
\multicolumn{1}{|c}{\citeauthor{liu2021statistically}} &
  time-oblivious &
  unsupervised &
  \begin{tabular}[c]{@{}c@{}}BERT-base,\\ LatinBERT~\citeyear{latinbert}\end{tabular} &
  \begin{tabular}[c]{@{}c@{}}multilingual,\\ monolingual\end{tabular} &
  domain-adaptation &
  last four &
  sum &
  - &
  CD &
  \multicolumn{1}{c|}{\begin{tabular}[c]{@{}c@{}}English,\\ German,\\ Latin,\\ Swedish\end{tabular}} \\ \hline
\multicolumn{1}{|c}{\citeauthor{giulianelli2022fire}} &
  time-oblivious &
  unsupervised &
  XLM-R-base &
  multilingual &
  domain-adaptation &
  all &
  average &
  - &
  \begin{tabular}[c]{@{}c@{}}APD,\\ PRT\end{tabular} &
  \multicolumn{1}{c|}{\begin{tabular}[c]{@{}c@{}}English,\\ German,\\ Italian,\\ Latin,\\ Norwegian,\\ Russian,\\ Swedish\end{tabular}} \\ \hline
\multicolumn{1}{|c}{\citeauthor{laicher2020clims}} &
  time-oblivious &
  unsupervised &
  mBERT-base &
  multilingual &
  pre-trained &
  \begin{tabular}[c]{@{}c@{}}all, \\ last four\end{tabular} &
  average &
  - &
  APD &
  \multicolumn{1}{c|}{Italian} \\ \hline
\multicolumn{1}{|c}{\citeauthor{qiu2022histbert}} &
  time-oblivious &
  unsupervised &
  BERT-base &
  monolingual &
  \begin{tabular}[c]{@{}c@{}}domain-adaptation\\ pre-trained\end{tabular} &
  last four &
  sum &
  - &
  CD &
  \multicolumn{1}{c|}{English} \\ \hline
\multicolumn{1}{|c}{\citeauthor{periti2022what}} &
  time-oblivious &
  unsupervised &
  \begin{tabular}[c]{@{}c@{}}BERT-base\\ mBERT-base\end{tabular} &
  \begin{tabular}[c]{@{}c@{}}monolingual,\\ multilingual\end{tabular} &
  pre-trained &
  last four &
  sum &
  - &
  \begin{tabular}[c]{@{}c@{}}CD,\\ DIV\end{tabular} &
  \multicolumn{1}{c|}{\begin{tabular}[c]{@{}c@{}}English,\\ Latin\end{tabular}} \\ \hline
\multicolumn{1}{|c}{\citeauthor{montariol2021scalable}} &
  time-oblivious &
  unsupervised &
  \begin{tabular}[c]{@{}c@{}}BERT-base\\ mBERT-base\end{tabular} &
  \begin{tabular}[c]{@{}c@{}}monolingual,\\ multilingual\end{tabular} &
  domain-adaptation &
  last four &
  sum &
  - &
  CD &
  \multicolumn{1}{c|}{\begin{tabular}[c]{@{}c@{}}English,\\ German,\\ Latin,\\ Swedish\end{tabular}} \\ \hline
  \multicolumn{1}{|c}{\citeauthor{cassotti2023xl}} & time-oblivious & supervised & \begin{tabular}[c]{@{}c@{}}XLM-R-large \\ (XL-LEXEME)\end{tabular} & \begin{tabular}[c]{@{}c@{}}multilingual\end{tabular} & fine-tuned & - & - & \begin{tabular}[c]{@{}c@{}}-\end{tabular} & \begin{tabular}[c]{@{}c@{}}APD\end{tabular} & \multicolumn{1}{c|}{\begin{tabular}[c]{@{}c@{}}English,\\ German,\\ Latin,\\ Swedish,\\ Russian\end{tabular}}\\ \hline
\end{tabular}%
}
\caption{Summary view of form-based approaches. Missing information is denoted with a dash}
\label{tab:form}
\end{table}

According to Table~\ref{tab:form}, we note that most form-based approaches are time-oblivious. A few time-aware approaches have been recently appeared and they are all characterised by the adoption of a specific fine-tuning operation to inject time information into the model. All the papers leverage unsupervised learning modalities with the exception of~\cite{arefyev2021deep}. The aggregation stage is mostly based on averaging, while clustering is only enforced in~\cite{beck2020diasense} where a cluster represents the dominant sense of the word $w$. In particular, in~\cite{beck2020diasense}, a word is considered as changing when clustering the embeddings $\Phi_w^1$ and $\Phi_w^2$ via K-means with $k=2$ generates two groups where one of the two clusters contains at least 90\% of the embeddings from one corpus only ($C_w^1$ or $C_w^2$). 

In form-based approaches, the following shift functions are proposed for measuring the semantic shift $s_w$.\\

\textbf{Cosine distance (CD).}
The shift $s_w$ is measured as the \textit{cosine distance} (CD) between the word prototypes $\mu_w^1, \mu_w^2$ as follows:
\begin{equation}
   CD(\mu_w^1, \mu_w^2) = 1-CS(\mu_w^1, \mu_w^2)
\end{equation}
where $CS$ is the \textit{cosine similarity} between the prototypes. Intuitively, the greater the $CD(\mu_w^1, \mu_w^2)$, the greater the shift in the dominant sense of $w$. 

Typically, the prototypes $\mu_w^1$ and $\mu_w^2$ are determined through aggregation by averaging over $\Phi_w^1$ and $\Phi_w^2$, respectively (e.g.,~\cite{martinc2020leveraging}). As a difference, in~\cite{horn2021exploring}, the prototype embedding $\mu_w^2$ at time step $t=2$ is computed by updating the prototype embedding $\mu_w^1$ at time step $t=1$ through a weighted running average~(e.g.,~\cite{finch2009incremental}).  

In~\cite{martinc2020leveraging}, the CD metric is employed in a multilingual experiment where the shift is measured across a diachronic corpus with texts of different languages. This is the only example of cross-language shift detection.

CD is also used in time-aware approaches. The integration of extra-linguistic information into word embeddings, such as time and social space, has been proposed in previous work based on static models~\citep{maja2018dynamic,ziqian2018biased}. Recently, this integration has been also applied to contextualised embeddings~\citep{huang2019neural,rottger2021temporal}. In~\cite{hofmann2021dynamic}, a pre-trained model is fine-tuned to encapsulate time and social space in the embedding model~\citep{hofmann2021dynamic}. Then, the shift $s_w$ is assessed by computing the CD between embeddings generated by the original pre-trained model and the embeddings generated by the time-aware, fine-tuned model.  In particular, in~\cite{zhou2020temporalteller}, a \textit{temporal referencing} mechanism is adopted to encode time-awareness into a pre-trained model. Temporal referencing is a pre-processing step of the documents that tags each occurrence of $w$ in $C_w^1$ and $C_w^2$ with a special marker denoting the corpus/time in which it  appears~\citep{ferrari2017detecting,dubossarsky2019time}. 
%For example, the word \verb|mouse| in the sequence \textit{a computer mouse[2000] is the device that moves the pointer or cursor on computer screen}, is tagged with the temporal marker \verb|[2000]|.
The embeddings of a tagged word are learned by fine-tuning the model for domain-adaptation. In this case, $s_w$ is assessed by computing the CD between $\mu_{w[1]}^1$ and $\mu_{w[2]}^1$, where $w[i]$ denotes $w$ with the temporal marker $t_i$. Similarly to~\cite{zhou2020temporalteller}, a time-aware approach is proposed in~\cite{rosin2021time} where a time marker is added to documents instead of words and the model is fine-tuned to predict the injected time information.
%For example, the tag \verb|[2022]| is introduced at the beginning of the sentence \textit{[2022] Joe Biden is the President of the United States}.
As an alternative, in~\cite{rosin2022temporal}, a {\em temporal attention} mechanism is adopted to generate the embeddings $\Phi_w^1$ and $\Phi_w^2$ for calculating CD. \\

\textbf{Inverted similarity over word prototype (PRT).}
This measure is proposed as an alternative to CD for improving the  effectiveness of the shift detection~\citep{kutuzov2020uio}. The \textit{inverted similarity over word prototypes} (PRT) measure is defined as:
\begin{equation}
    PRT(\mu_w^1, \mu_w^2) = \frac{1}{CS(\mu_w^1, \mu_w^2)} \ .
\end{equation}

\textbf{Time-diff (TD).}
This measure is designed for time-aware approaches and it works on analysing the change of polysemy of a word along time. It is based on the model capability to predict the time of a document and it calculates the shift $s_w$ by considering the probability distribution of the predicted times~\citep{rosin2021time}. Intuitively, a uniform distribution means that the association document-time is not strong enough to clearly entail a shift. Instead, a non-uniform distribution means that there is an evidence to predict the time of a document. Consider a document $d_w$, let $p_t(d_w)$ be the probability of $d_w$ to belong to the time $t$. The function {\em time diff} (TD) is defined as the average difference of the predicted time probabilities:
\begin{equation}
    TD(C_w^1, C_w^2) = \frac{1}{|C_w|} \sum_{d_w^1 \in C_w^1, d_w^2 \in C_w^2} |p_1(d_w^1) - p_2(d_w^2)| \ .
\end{equation}
The experiments performed in~\cite{rosin2021time} show that TD outperforms CD on short-term semantic shift. On the contrary, CD outperforms TD over long-term semantic shift. We argue that the time-diff measure is more effective on long-term periods since major differences in writing style emerge and the prediction of document-time associations is more reliable.\\  

\textbf{Average pairwise distance (APD).}
This measure exploits the variance of the contextualised representations $\Phi_w^1$, $\Phi_w^2$ to compute the semantic shift assessment (i.e., variance on the word polysemy) . As a difference with the previous measures, APD directly works on word-occurrence embeddings without requiring any aggregation stage, namely clustering nor averaging. The \textit{average pairwise distance} (APD) is defined as follows:
\begin{equation}
    APD(\Phi_w^1, \Phi_w^2) = \frac{1}{|\Phi_w^2||\Phi_w^2|} \ \cdot %\\
    \sum_{e_{w,i}^1\in \ \Phi_w^1, \ e_{w,i}^2 \in \ \Phi_w^2} d(e_{w,i}^1, e_{w,i}^2) \ ,
\end{equation}
where $d$ is an arbitrary distance measure (e.g., cosine distance, Euclidean distance, Canberra distance). According to the experiments performed in~\cite{giulianelli2020analysing}, APD better performs when the Euclidean distance is employed as $d$. In~\cite{keidar2022slangvolution}, APD is used over the embeddings $\Phi_w^1$ and $\Phi_w^2$ by applying a dimensionality reduction through the Principal Component Analysis (PCA). In~\cite{keidar2022slangvolution}, experiments on both slang and non-slang words are performed through causal analysis to study how distributional factors (e.g., polysemy, frequency shift) influence the shift $s_w$. The results show that slang words experience fewer semantic shifts than non-slang words.

In~\cite{kudisov2022bos}, lexical substitutes are used to assess $s_w$. Given a word $w$, lexical substitutes of $w$ are those words that can replace $w$ in a text fragment without introducing grammatical errors or significantly changing its meaning. 
%For example, suitable substitutes for the word \verb|fly| in the sentence \textit{a noisy fly sat on my shoulder} are \verb|bug|, \verb|beetle|, or \verb|butterfly|; while suitable substitutes in the sentence \textit{we will fly to London} are \verb|walk|, \verb|run|, or \verb|bike|. 
A set of lexical substitutes is generated by leveraging a masked language model (e.g., XLM-R). In this case, the word representations $\Phi_w^1$, and $\Phi_w^2$ are defined as the bag-of-words vectors computed over the substitutes (i.e., \textit{bag-of-substitutes}) through Tf-Idf. Then, APD is finally computed over $\Phi_w^1$, and $\Phi_w^2$ to assess $s_w$. 

APD is also used in a time-aware approach described in~\cite{pomsl2020circe}, where a pre-trained BERT model is fine-tuned to predict the time period of a sentence. APD is finally used to measure the shift between the embeddings extracted from the fine-tuned model.

In~\cite{arefyev2021deep,cassotti2023xl}, ADP is employed to measure the shift $s_w$ over the embeddings $\Phi_w^1$ and $\Phi_w^2$ extracted from a supervised Word-in-Context model (WiC)~\citep{pilehvar2019wic}. This model is trained to reproduce the behavior of human annotators when they are asked to evaluate the similarity of the meaning of a word $w$ in a pair of given sentences from $C_w^1$ and $C_w^2$, respectively. The embeddings $\Phi_w^1$ and $\Phi_w^2$ are extracted from the trained WiC model for calculating the final APD measure. 

\textbf{Average of average inner distances (APD-OLD/NEW).}
The APD-OLD/NEW measure is presented in~\cite{laicher2021explaining} as an extension of APD and it estimates the shift $s_w$ as the average degree of polysemy of $w$ in the corpora $C_w^1$ and $C_w^2$, respectively. The \textit{average of average inner distances} (APD-OLD/NEW) is defined as:
\begin{equation}
    APD\text{-}OLD/NEW(\Phi_w^1, \Phi_w^2) = \frac{AID(\Phi_w^1) + AID(\Phi_w^2)}{2} \ .
\end{equation}
where AID is the \textit{average inner distance} and it measures the degree of polysemy of $w$ in a specific time frame by relying on the APD measure, namely $AID(\Phi_w^1) = APD(\Phi_w^1, \Phi_w^1)$ and $AID(\Phi_w^2) = APD(\Phi_w^2, \Phi_w^2)$, respectively.\\ 

\textbf{Hausdorff distance (HD).}
The shift $s_w$ is measured as the \textit{Hausdorff distance} (HD) between the word embeddings $\Phi_w^1$ and $\Phi_w^2$. 
% Similarly to APD, HD directly works on word-occurrence embeddings without requiring any aggregation stage. 
HD relies on the Euclidean distance $d$ to measure the difference between the embeddings of $w$ in $C_w^1$ and $C_w^2$ and it returns the greatest of all the distances $d$ from one embedding $e_w^1 \in \Phi_w^1$ to the closest embedding $e_w^2 \in \Phi_w^2$, or vice-versa. The HD measure is defined as follows: 
\begin{equation}
    HD(\Phi_w^1, \Phi_w^2) = \max \left(\sup_{e_w^1 \in \Phi_w^1} \inf_{e_w^2 \in \Phi_w^2} d(e_w^1, e_w^2), \sup_{e_w^2 \in \Phi_w^2} \inf_{e_w^1 \in \Phi_w^1} d(e_w^2, e_w^1) \right) \ .
\end{equation}
The experiments performed in~\cite{wang2020university} show that HD is sensitive to outliers since it is based on infimum and supremum, thus an outlier embedding may largely affect the final $s_w$ value. \\

\textbf{Difference between token embedding diversities (DIV).}
Similar to APD, this measure assesses the shift $s_w$ by exploiting the variance of the contextualised representation $\Phi_w^1$ and $\Phi_w^2$. As a difference with APD, the \textit{difference between token embedding diversities} (DIV) leverage a coefficient of variation calculated as the average of the cosine distances $d$ between the embeddings $\Phi_w^1$ and $\Phi_w^2$, and their prototypical embeddings $\mu_w^1$ and $\mu_w^2$, respectively~\cite{kutuzov2020distributional}. The intuition is that when $w$ is used in just one sense, its embeddings tend to be close to each other yielding a low coefficient of variation. On the opposite, when $w$ is used many different senses, its embeddings are distant to each other yielding to a high coefficient of variation. DIV is defined as the absolute difference between the coefficient of variation in $C_w^1$ and $C_w^2$:
\begin{equation}
    DIV(\Phi_w^1, \Phi_w^2) = \left| \frac{\sum_{e_w^1 \in \Phi_w^1} d(e_w^1, \mu_w^1)}{|\Phi_w^1|} - \frac{\sum_{e_w^2 \in \Phi_w^2} d(e_w^2, \mu_w^2)}{|\Phi_w^2|}\right|
\end{equation}
In~\cite{kutuzov2020distributional}, the experiments show that when the coefficient of variation is low, the prototypical embeddings $\mu_w^1$ and $\mu_w^2$ successfully represent the meanings of the given word $w$. On the opposite, when the coefficient of variation is high, the prototypical embeddings $\mu_w^1$ and $\mu_w^2$ do not provide a relevant representation of the $w$ meanings.

\subsection{Sense-based approaches}\label{sec:sense}

\begin{table}[!ht]
\centering
\resizebox{\textwidth}{!}{%
\begin{tabular}{ccccccccccc}
\textbf{Ref.} & \textbf{\begin{tabular}[c]{@{}c@{}}Time\\ awareness\end{tabular}} & \textbf{\begin{tabular}[c]{@{}c@{}}Learning\\ modality\end{tabular}} & \textbf{\begin{tabular}[c]{@{}c@{}}Language\\ model\end{tabular}} & \textbf{\begin{tabular}[c]{@{}c@{}}Training\\ language\end{tabular}} & \textbf{\begin{tabular}[c]{@{}c@{}}Type of \\ training\end{tabular}} & \textbf{Layer} & \textbf{\begin{tabular}[c]{@{}c@{}}Layer \\ aggregation\end{tabular}} & \textbf{\begin{tabular}[c]{@{}c@{}}Clustering\\ algorithm\end{tabular}} & \textbf{\begin{tabular}[c]{@{}c@{}}Shift \\ function\end{tabular}} & \textbf{\begin{tabular}[c]{@{}c@{}}Corpus \\ language\end{tabular}} \\ \hline
\multicolumn{1}{|c}{\citeauthor{hu2019diachronic}} & time-oblivious & supervised & BERT-base & monolingual & pre-trained & last & - & - & MNS & \multicolumn{1}{c|}{English} \\ \hline
\multicolumn{1}{|c}{\citeauthor{rachinskiy2021zeroshot}} & time-oblivious & supervised & XLM­-R-base & multilingual & \begin{tabular}[c]{@{}c@{}}fine-tuned,\\ pre-trained\end{tabular} & - & - & - & APD & \multicolumn{1}{c|}{Russian} \\ \hline
\multicolumn{1}{|c}{\citeauthor{rachinskiy2022glossreader}} & time-oblivious & supervised & XLM­-R-base & multilingual & \begin{tabular}[c]{@{}c@{}}fine-tuned,\\ pre-trained\end{tabular} & last & - & - & \begin{tabular}[c]{@{}c@{}}APD,\\ JSD\end{tabular} & \multicolumn{1}{c|}{Spanish} \\ \hline
\multicolumn{1}{|c}{\citeauthor{periti2022what}} & time-oblivious & unsupervised & \begin{tabular}[c]{@{}c@{}}BERT-base,\\ mBERT-base\end{tabular} & \begin{tabular}[c]{@{}c@{}}monolingual,\\ multilingual\end{tabular} & pre-trained & last four & sum & \begin{tabular}[c]{@{}c@{}}AP, \\ APP,\\ IAPNA\end{tabular} & \begin{tabular}[c]{@{}c@{}}JSD,\\ PDIS,\\ PDIV\end{tabular} & \multicolumn{1}{c|}{\begin{tabular}[c]{@{}c@{}}English,\\ Latin\end{tabular}} \\ \hline
\multicolumn{1}{|c}{\citeauthor{montariol2021scalable}} & time-oblivious & unsupervised & \begin{tabular}[c]{@{}c@{}}BERT-base,\\ mBERT-base\end{tabular} & \begin{tabular}[c]{@{}c@{}}monolingual,\\ multilingual\end{tabular} & domain-adaptation & last four & sum & \begin{tabular}[c]{@{}c@{}}K-Means,\\ AP\end{tabular} & \begin{tabular}[c]{@{}c@{}}JSD,\\ WD\end{tabular} & \multicolumn{1}{c|}{\begin{tabular}[c]{@{}c@{}}English,\\ German,\\ Latin,\\ Swedish\end{tabular}} \\ \hline
\multicolumn{1}{|c}{\citeauthor{rodina2020elmo}} & time-oblivious & unsupervised & \begin{tabular}[c]{@{}c@{}}mBERT-base,\\ ELMo\end{tabular} & \begin{tabular}[c]{@{}c@{}}monolingual,\\ multilingual\end{tabular} & domain-adaptation & last & - & \begin{tabular}[c]{@{}c@{}}K-Means,\\ AP\end{tabular} & \begin{tabular}[c]{@{}c@{}}JSD\\ MS\end{tabular} & \multicolumn{1}{c|}{Russian} \\ \hline
\multicolumn{1}{|c}{\citeauthor{kanjirangat2020sst}} & time-oblivious & unsupervised & mBERT-base & multilingual & pre-trained & last four & concatenation & K-Means & \begin{tabular}[c]{@{}c@{}}CSC,\\ JSD\end{tabular} & \multicolumn{1}{c|}{\begin{tabular}[c]{@{}c@{}}English,\\ German,\\ Latin,\\ Swedish\end{tabular}} \\ \hline
\multicolumn{1}{|c}{\citeauthor{giulianelli2020analysing}} & time-oblivious & unsupervised & BERT-base & monolingual & pre-trained & all & sum & K-Means & \begin{tabular}[c]{@{}c@{}}ED,\\ JSD\end{tabular} & \multicolumn{1}{c|}{English} \\ \hline
\multicolumn{1}{|c}{\citeauthor{arefyev2020bos}} & time-oblivious & unsupervised & XLM-R-base & multilingual & domain-adaptation & - & - & AGG & CDCD & \multicolumn{1}{c|}{\begin{tabular}[c]{@{}c@{}}English,\\ German,\\ Latin,\\ Swedish\end{tabular}} \\ \hline
\multicolumn{1}{|c}{\citeauthor{kashleva2022hse}} & time-oblivious & unsupervised & BERT-base & monolingual & domain-adaptation & all & sum & K-Means & APDP & \multicolumn{1}{c|}{Spanish} \\ \hline
\multicolumn{1}{|c}{\citeauthor{martinc2020discovery}} & time-oblivious & unsupervised & \begin{tabular}[c]{@{}c@{}}BERT-base,\\ mBERT-base\end{tabular} & \begin{tabular}[c]{@{}c@{}}monolingual,\\ multilingual\end{tabular} & domain-adaptation & last four & sum & \begin{tabular}[c]{@{}c@{}}K-Means,\\ AP\end{tabular} & JSD & \multicolumn{1}{c|}{\begin{tabular}[c]{@{}c@{}}English,\\ German,\\ Latin,\\ Swedish\end{tabular}} \\ \hline
\multicolumn{1}{|c}{\citeauthor{kutuzov2020uio}} & time-oblivious & unsupervised & \begin{tabular}[c]{@{}c@{}}BERT-base,\\ ELMo,\\ mBERT-base\end{tabular} & \begin{tabular}[c]{@{}c@{}}monolingual,\\ multilingual\end{tabular} & \begin{tabular}[c]{@{}c@{}}domain-adaptation,\\ incremental domain-adaptation,\\ pre-trained\end{tabular} & \begin{tabular}[c]{@{}c@{}}all,\\ last,\\ last four\end{tabular} & average & AP & JSD & \multicolumn{1}{c|}{\begin{tabular}[c]{@{}c@{}}English,\\ German,\\ Latin,\\ Swedish\end{tabular}} \\ \hline
\multicolumn{1}{|c}{\citeauthor{giulianelli2022fire}} & time-oblivious & unsupervised & XLM-R-base & multilingual & domain-adaptation & all & average & AP & JSD & \multicolumn{1}{c|}{\begin{tabular}[c]{@{}c@{}}English,\\ German,\\ Italian,\\ Latin,\\ Norwegian,\\ Russian,\\ Swedish\end{tabular}} \\ \hline
\multicolumn{1}{|c}{\citeauthor{wang2020university}} & time-oblivious & unsupervised & mBERT-base & multilingual & domain-adaptation & last & - & \begin{tabular}[c]{@{}c@{}}GMMs,\\ K-Means\end{tabular} & JSD & \multicolumn{1}{c|}{Italian} \\ \hline
\multicolumn{1}{|c}{\citeauthor{keidar2022slangvolution}} & time-oblivious & unsupervised & RoBERTa-base & monolingual & domain-adaptation & \begin{tabular}[c]{@{}c@{}}all, \\ first,\\ last\end{tabular} & sum & \begin{tabular}[c]{@{}c@{}}AP,\\ K-Means,\\ GMMs\end{tabular} & \begin{tabular}[c]{@{}c@{}}ED,\\ JSD\end{tabular} & \multicolumn{1}{c|}{English} \\ \hline
\multicolumn{1}{|c}{\citeauthor{karnysheva2020tue}} & time-oblivious & unsupervised & \begin{tabular}[c]{@{}c@{}}ELMo,\\ mELMo\end{tabular} & \begin{tabular}[c]{@{}c@{}}monolingual,\\ multilingual\end{tabular} & pre-trained & all & - & \begin{tabular}[c]{@{}c@{}}K-Means,\\ DBSCAN\end{tabular} & JSD & \multicolumn{1}{c|}{\begin{tabular}[c]{@{}c@{}}English,\\ German,\\ Latin,\\ Swedish\end{tabular}} \\ \hline
\multicolumn{1}{|c}{\citeauthor{cuba2020sensecluster}} & time-oblivious & unsupervised & XLM-R-base & multilingual & pre-trained & last & - & K-Means & JSD & \multicolumn{1}{c|}{\begin{tabular}[c]{@{}c@{}}English,\\ German,\\ Latin,\\ Swedish\end{tabular}} \\ \hline
\multicolumn{1}{|c}{\citeauthor{rother2020cmce}} & time-oblivious & unsupervised & \begin{tabular}[c]{@{}c@{}}mBERT-base,\\ XLM-R-base\end{tabular} & multilingual & fine-tuned & last & - & \begin{tabular}[c]{@{}c@{}}BIRCH,\\ DBSCAN,\\ GMMs,\\ HDBSCAN\end{tabular} & JSD & \multicolumn{1}{c|}{\begin{tabular}[c]{@{}c@{}}English,\\ German,\\ Latin,\\ Swedish\end{tabular}} \\ \hline
\multicolumn{1}{|c}{\citeauthor{periti2023studying}} & time-oblivious & unsupervised & \begin{tabular}[c]{@{}c@{}}BERT-base,\\mBERT-base,\\ XLM-R-base\end{tabular} & \begin{tabular}[c]{@{}c@{}}monolingual,\\ multilingual\end{tabular} & pre-trained & last & - & \begin{tabular}[c]{@{}c@{}}AP,\\ APP\end{tabular} & \begin{tabular}[c]{@{}c@{}}JSD,\\ APDP\end{tabular} & \multicolumn{1}{c|}{\begin{tabular}[c]{@{}c@{}}English,\\ German,\\ Latin,\\ Swedish,\\ Spanish,\\ Russian,\\Italian\end{tabular}} \\ \hline
\end{tabular}%
}
\caption{Summary view of sense-based approaches. Missing information is denoted with a dash}
\label{tab:sense}
\end{table}

According to Table~\ref{tab:sense}, we note that all the sense-based approaches are time-oblivious and that fine-tuning is sometimes adopted, but mainly for domain-adaptation purposes. Most papers leverage unsupervised learning modalities. Only a few exceptions employ a lexicographic supervision (i.e.,~\cite{hu2019diachronic,rachinskiy2021zeroshot,rachinskiy2022glossreader}). As a difference with form-based, sense-based approaches usually enforce clustering in the aggregation stage. The aggregation by averaging is only exploited in~\cite{periti2022what,hu2019diachronic,montariol2021scalable} where sense prototypes are computed on top of the results of a clustering operation. 

When clustering is adopted, the function $f$ that calculates the shift $s_w$ can be directly defined over the embeddings $\Phi_w^1$ and $\Phi_w^2$. As an alternative, the function $f$ can be defined over the distribution of the embeddings in the resulting clusters (i.e., {\em cluster distribution}). In this case, as a result of the clustering operation, a counting function $c$ is used 
%a mapping function $m$ is used to associate each embedding to the corresponding cluster label. A counting function $c$ is defined over $m$ 
to determine two cluster distributions $p_w^1$ and $p_w^2$ that represent the normalised number of embeddings in the cluster partitions $\phi_{w,i}^1$ and $\phi_{w,i}^2$, respectively (see Section~\ref{sec:problem}). The $i$-th value $p_{w,i}^j$ in $p_w^j$ (with $j \in \{1, 2\}$) represents the number of embeddings of $\phi_{w,i}^j$ in the $i$-th cluster, namely:
\begin{equation}
    p_{w,i}^j = \frac{|\phi_{w,i}^j|}{|\Phi_{w}^j|} \ .
\end{equation}
Finally, the function $f$ is defined as a compound function $f = g \circ c$, where the result of the $c$ function is exploited by a shift function $g$ which works on the cluster distributions $p_w^1$ and $p_w^2$. 

In sense-based approaches, the following shift functions are proposed for measuring the semantic shift $s_w$.\\

\textbf{Maximum novelty score (MNS).}
This measure exploits the cluster distributions $p_w^1$ and $p_w^2$ by leveraging the idea that the higher is the ratio between the number of embeddings $\Phi_w^1$ and $\Phi_w^2$ in a cluster, the higher is the semantic shift of the considered word $w$. The \textit{maximum novelty score} (MNS) is defined as:
\begin{equation}
    MNS(p_{w}^1, p_{w}^2) = \max \{NS(p_{w,1}^1, p_{w,1}^2), ..., NS(p_{w,k}^2, p_{w,k}^2)\} \ ,
\end{equation}
where $NS(p_{w,i}^1, p_{w,i}^2) = p_{w,i}^1/p_{w,i}^2$ is the \textit{novelty score} proposed in~\cite{cook2014novel}, and $k$ is the number of clusters produced as a result of the aggregation stage.

In~\cite{hu2019diachronic}, MNS is employed as a shift measure in a supervised learning approach. In particular, a lexicographic supervision (i.e., the Oxford English dictionary) is employed to provide the meanings of the target word $w$. Each word occurrence in $\Phi_w^1$ and $\Phi_w^2$ is associated with the closest meaning of the dictionary according to the cosine distance. As a result, for each word/dictionary meaning, a cluster of word embeddings is defined and MNS is exploited to calculate the overall shift.\\

\textbf{Maximum square (MS).} 
This measure is an alternative to MNS to assess the shift of $s_w$. The intuition of MS is that slight changes in cluster distributions $p_w^1$ and $p_w^2$ may occur due to noise and do not represent a real semantic shift~\citep{rodina2020elmo}. The \textit{maximum square} (MS) aims at identifying strong changes in the cluster distributions. As a difference with MNS, the square difference between $p_{w,i}^1$ and $p_{w,i}^2$ is used to capture the degree of shift instead of the novelty score (NS):
\begin{equation}
    MS(p_w^1, p_w^2) = \max_i \left(p_{w,i}^1-p_{w,i}^2\right)^2
\end{equation}

\textbf{Jensen-Shannon divergence (JSD).} 
This measure extends the Kullback-Leibler (KL) divergence, which calculates how one probability distribution is different from another. The \textit{Jensen-Shannon divergence} (JSD) calculates the shift $s_w$ as the symmetrical KL score of the cluster distributions $p_w^1$ from $p_w^2$, namely:
\begin{equation}
    JSD(p_w^1, p_w^2) = \frac{1}{2}\left(KL(p_w^1||M) + KL(p_w^2||M)\right)
    \ ,
\end{equation}
where KL is the Kullback-Leibler divergence and $M=(p_w^1+p_w^2)/2$.

JSD is also used in approaches where aggregation by clustering is performed separately over the embeddings $\Phi_w^1$ and $\Phi_w^2$~\citep{kanjirangat2020sst}. As a result, the clusters need to be aligned to determine the distributions $p_w^1$ and $p_w^2$ before the JSD calculation. As a difference with~\cite{kanjirangat2020sst}, an evolutionary clustering algorithm is employed in~\cite{periti2022what} to apply the JSD measure without requiring any alignment step over the resulting clusters.

As a final remark, JSD can be employed to measure the shift $s_w$ over more than two time periods. However, the experiments in~\cite{giulianelli2020analysing} show that the JSD effectiveness over a single time period outperforms the version over more time periods since JSD is insensitive to the order of the temporal intervals. \\ 

\textbf{Coefficient of semantic change (CSC).} 
This measures is proposed as an alternative to JSD where the difference over the weighted number of elements in $\phi_{w,i}^1$ and $\phi_{w,i}^2$ for each cluster $i$ is employed to replace KL in measuring the shift~\citep{kanjirangat2020sst}. The \textit{coefficient of semantic change} (CSC) is defined as follows:
\begin{equation}
    CSC(p_w^1, p_w^2) = \frac{1}{P_w^1 \cdot P_w^2} \sum^K_{k=1} |P_w^2 \cdot p_{w,k}^1 - P_w^1 \cdot p_{w,k}^2| \ ,
\end{equation}
where $P_w^j = \sum^k_{i=1} p_{w,i}^j$ is the weight of each cluster distribution and $k$ is the number of clusters.\\

\textbf{Cosine distance between cluster distributions (CDCD).}
As a further alternative of JSD, this measure assess the shift $s_w$ by considering the cluster distributions $p_w^1$ and $p_w^2$ as vectors and by applying the cosine distance over them to assess the semantic shift $s_w$. The \textit{cosine distance between cluster distributions} (CDCD) is defined as follows:
\begin{equation}
    CDCD(p_w^1, p_w^2) = 1- \frac{p_w^1 \cdot p_w^2}{\|p_w^1\| \times \|p_w^2\|}
\end{equation}
In~\cite{arefyev2020bos}, CDCD is calculated between the cluster distributions $p_w^1$ and $p_w^2$ obtained by enforcing clustering over bag-of-substitutes (see the description of~\cite{arefyev2020bos} in Section~\ref{sec:form}).\\

\textbf{Entropy difference (ED).} 
This measure is based on the idea that the higher is the uncertainty in the interpretation of a word occurrence due to the $w$ polysemy in $C_w^1$ and $C_w^2$, the higher is the semantic shift $s_w$. The intuition is that high values of ED are associated with the broadening of a word’s interpretation, while negative values indicate a narrowing interpretation~\citep{giulianelli2020analysing}. The \textit{entropy difference} (ED) is defined as follows:
\begin{equation}
    ED(p_{w}^1, p_{w}^2) = \eta(p_{w}^1) - \eta(p_{w}^2) \ ,
\end{equation} 
where $\eta(p_{w}^j)$ is the degree of polysemy of $w$ in the corpus $C_j$, which is calculated as the normalised entropy of its cluster distribution $p_{w}^j$:
\begin{equation*}
    \eta(p_{w}^j) = \log_{K_w} \left(\prod_{k=1}^{K_w} {p_{w,i}^j}^{-p_{w,i}^j} \right) \ .
\end{equation*}
As shown in~\cite{giulianelli2020analysing}, ED is not capable of properly assessing $s_w$ when new usage types of $w$ emerge, while old ones become obsolescent at the same time, since it may lead to no entropy reduction. \\

\textbf{Cosine distance between word prototypes (PDIS).}
This measure is presented in~\cite{periti2022what} as an extension of the CD measure adopted by form-oriented approaches. The idea of PDIS is that the aggregation by averaging over cluster prototypes can be employed to produce summary descriptions of the cluster contents (i.e., {\em semantic prototypes}). The \textit{cosine distance between word prototypes}  (PDIS) is defined as the CD between $\bar{c}_{w}^1$, $\bar{c}_{w}^2$, that is:
\begin{equation}
    PDIS(\bar{c}_{w}^1, \bar{c}_{w}^2) = 1- \frac{\bar{c}_{w}^1 \cdot \bar{c}_{w}^2}{\| \bar{c}_{w}^1\| \times \|\bar{c}_{w}^2\|}
\end{equation}
where $\bar{c}_{w}^1$ and $\bar{c}_{w}^2$ are semantic prototypes defined as the average embeddings of all the sense prototypes $c_{w,i}^1$ and $c_{w,i}^2$, respectively.\\

\textbf{Difference between prototype embedding diversities (PDIV).}
This measure is presented in~\cite{periti2022what} as an extension of the DIV measure adopted by form-oriented approaches. PDIV leverages the same intuition of PDIS, namely the semantic prototypes can be employed to calculate the coefficient of ambiguity of $w$ by measuring the difference between a semantic prototype $\bar{c}_{w}^j$ and each sense prototype $c_{w,i}^j$. The \textit{difference between prototype embedding diversities} (PDIV) is defined as the absolute difference between these ambiguity coefficients:
\begin{equation}
    PDIV(\Psi_w^1, \Psi_w^2) = \left|\frac{\sum_{c_{w,k}^1 \in \Psi_w^1} \ d(c_{w,k}^1, \bar{c}_{w}^1)}{|\Psi_w^1|}\right. 
    - \left.\frac{\sum_{c_{w,k}^2 \in \Psi_w^2} \ d(c_{w,k}^2, \bar{c}_{w}^2)}{|\Psi_w^2|}\right| \ ,
\end{equation}
where $\Psi_w^1$ and $\Psi_w^2$ denote the set of sense prototypes of $c_{w,i}^1$ and $c_{w,i}^2$, respectively. \\

\textbf{Average pairwise distance (APD).}
In addition to form-based approaches (see Section~\ref{sec:form}), the APD measure is exploited to assess $s_w$ also in sense-based approaches. In~\cite{rachinskiy2021zeroshot,rachinskiy2022glossreader}, APD is applied to the contextualised embeddings $\Phi_w^1$ and $\Phi_w^2$ extracted from a fine-tuned XLM-R model. In particular, an English corpus is used to fine-tune the pre-trained model to select the most appropriate WordNet's definition for each word occurrence~\citep{blevins2020moving}. As a result of the fine-tuning, both WordNet's definitions and word occurrences are embedded in the same vector space and the meaning of any word occurrence can be induced by selecting the closest definition in the vector space. In~\cite{rachinskiy2021zeroshot}, the zero-shot, cross-lingual transferability property of XLM-R is exploited to obtain word representations for Russian language and APD is finally applied~\citep{ming2008importance,choi2021analyzing}. The authors of~\cite{rachinskiy2021zeroshot} claim that the approach is useful to overstep the lack of lexicographic supervision for low-resource languages and that most concept definitions in English also hold in other languages, such as Russian. However, this claim is not completely satisfied, since some words can drastically change their meaning across languages. For example, the Russian word "\foreignlanguage{russian}{пионер}"  (pioneer, scout) is strongly connected to the Communist ideology in the Soviet Period, but it isn't in the English language.\\

\textbf{Average pairwise distance between sense prototypes (APDP).}
This measure is an extension of APD and it considers all the pairs of sense prototypes $c_{w,i}^1$ and $c_{w,i}^2$ instead of all the original embeddings in $\Phi_w^1$ and $\Phi_w^2$~\citep{kashleva2022hse}. The \textit{average pairwise distance between sense prototypes (APDP)} is defined as:
\begin{equation}
    APD(\Psi_w^1, \Psi_w^2) = \frac{1}{|\Psi_w^2||\Psi_w^2|} \ \cdot 
    \sum_{c_{w,k}^1\in \Psi_w^1, \ c_{w,k}^2 \in \Psi_w^2} d(c_{w,k}^1, c_{w,k}^2) \ 
\end{equation}

\textbf{Wassertein distance (WD).} 
This measure models the shift assessment as an {\em optimal transport problem} and it is exploited as an alternative to cluster alignment when aggregation by clustering is performed separately over the embeddings $\Phi_w^1$ and $\Phi_w^2$~\citep{montariol2021scalable}.  
WD quantifies the effort of re-configuring the cluster distribution of $p_w^1$ into $p_w^2$, namely minimising the cost of moving one unit of mass (i.e., a sense prototype) from $\Psi_w^1$ to $\Psi_w^2$. The \textit{Wassertein distance} (WD) is defined as:
\begin{equation}
    WD(p_w^1, p_w^2) =  \min_\gamma \sum_i^{k_1} \sum_j^{k_2} CD(c_{w,i}^1, c_{w,j}^2) \ \gamma_{c_{w,i}^1 \rightarrow c_{w,j}^2}
\end{equation}
\begin{equation*}
    \begin{aligned}
    \text{such that:} & & \gamma_{c_{w,i}^1 \rightarrow c_{w,j}^2} \ge 0 \ \ \ \ \ \ \\
    & & \sum_i \gamma_{c_{w,i}^1 \rightarrow c_{w,j}^2} = p_w^1\\
    & & \sum_j \gamma_{c_{w,i}^1 \rightarrow c_{w,j}^2} = p_w^2
\end{aligned}
\end{equation*}
where all $\gamma_{c_{w,i}^1 \rightarrow c_{w,j}^2}$ represents the (unknown) effort required to reconfigure the mass distribution $p_w^1$ into $p_w^2$; $k_1$ and $k_2$ are the number of clusters obtained by clustering $\Phi_w^1$ and $\Phi_w^2$, respectively; $CD$ is the cosine distance computed over the sense prototypes $c_{w,i}^1 \in \Psi_w^1$ and $c_{w,j}^2 \in \Psi_w^2$~\citep{bonneel2011displacement}. 

%temporaneo
%%The zero-shot cross lingual transferability is exploited also by Teodorescu et al.. In particular, they approach binary semantic change detection by using AMuSE~\cite{teodorescu2022ualberta}, a neural Word Sense Disambiguation system trained on manual annotations involving English WordNet senses, which thanks to its use of the multilingual XLM-R embeddings, it is also applicabale to other languages that are represented in BableNet~\cite{orlando2021amuse}.
%The assumption of their approach is that to establish if a word $w$ change in meaning it may be sufficient to determine if the set of senses for a target word remains the same from the old to the modern corpus. For each target word, they  compute its sense frequency distributions in both the old and modern corpora based on the output of the WSD system. If a sense is found in the modern corpus but is missing in the old corpus (or vice versa), a change is deemed to have occurred. Otherwise, a word has the same set of senses identified in both the old and modern corpus. For each sense, they compute the relative probability change. The resulting value is compared to a threshold tuned on the development data by maximizing F-score. 

\subsection{Ensemble-based approaches}\label{sec:ensemble}
In this section, we review the \app\ approaches that rely on an {\em ensemble mechanism}, namely the combination of two or more assessment functions to determine the semantic shift score. Ensembling can mean that more than one form- and/or sense-based measure is adopted in a given approach. Ensembling can also mean that a disciplined use of both static and contextualised embedding models is used. A final semantic shift score is then returned by the whole ensemble process. 

\begin{table}[!h]
\centering
\resizebox{0.98\textwidth}{!}{%
\begin{tabular}{ccccccccccc}
\textbf{Ref.} &
  \textbf{\begin{tabular}[c]{@{}c@{}}Time\\ awareness\end{tabular}} &
  \textbf{\begin{tabular}[c]{@{}c@{}}Learning\\ modality\end{tabular}} &
  \textbf{\begin{tabular}[c]{@{}c@{}}Language\\ model\end{tabular}} &
  \textbf{\begin{tabular}[c]{@{}c@{}}Training\\ language\end{tabular}} &
  \textbf{\begin{tabular}[c]{@{}c@{}}Type of \\ training\end{tabular}} &
  \textbf{Layer} &
  \textbf{\begin{tabular}[c]{@{}c@{}}Layer \\ aggregation\end{tabular}} &
  \textbf{\begin{tabular}[c]{@{}c@{}}Clustering\\ algorithm\end{tabular}} &
  \textbf{\begin{tabular}[c]{@{}c@{}}Shift \\ function\end{tabular}} &
  \textbf{\begin{tabular}[c]{@{}c@{}}Corpus \\ language\end{tabular}} \\ \hline
\multicolumn{1}{|c}{\citeauthor{pomsl2020circe}} &
  time-aware &
  unsupervised &
  \begin{tabular}[c]{@{}c@{}}BERT-base,\\ mBERT-base\end{tabular} &
  \begin{tabular}[c]{@{}c@{}}monolingual,\\ multilingual\end{tabular} &
  fine-tuned &
  last &
  - &
  - &
  APD &
  \multicolumn{1}{c|}{\begin{tabular}[c]{@{}c@{}}English,\\ German,\\ Latin,\\ Swedish\end{tabular}} \\ \hline
\multicolumn{1}{|c}{\citeauthor{teodorescu2022ualberta}} &
  time-oblivious &
  unsupervised &
  XLM-large &
  multilingual &
  trained &
  last four &
  sum &
  - &
  APD &
  \multicolumn{1}{c|}{Spanish} \\ \hline
\multicolumn{1}{|c}{\citeauthor{martinc2020discovery}} &
  time-oblivious &
  unsupervised &
  \begin{tabular}[c]{@{}c@{}}BERT-base,\\ mBERT-base\end{tabular} &
  \begin{tabular}[c]{@{}c@{}}monolingual,\\ multilingual\end{tabular} &
  domain-adaptation &
  last four &
  sum &
  AP &
  \begin{tabular}[c]{@{}c@{}}CD,\\ JSD\end{tabular} &
  \multicolumn{1}{c|}{\begin{tabular}[c]{@{}c@{}}English,\\ German,\\ Latin,\\ Swedish\end{tabular}} \\ \hline
\multicolumn{1}{|c}{\citeauthor{wang2020university}} &
  time-oblivious &
  unsupervised &
  mBERT-base &
  multilingual &
  pre-trained &
  last &
  - &
  \begin{tabular}[c]{@{}c@{}}GMMs,\\ K-Means\end{tabular} &
  \begin{tabular}[c]{@{}c@{}}APD,\\ HD,\\ JSD\end{tabular} &
  \multicolumn{1}{c|}{Italian} \\ \hline
\multicolumn{1}{|c}{\citeauthor{giulianelli2022fire}} &
  time-oblivious &
  unsupervised &
  XLM-R-base &
  multilingual &
  domain-adaptation &
  all &
  average &
  - &
  \begin{tabular}[c]{@{}c@{}}APD,\\ PRT\end{tabular} &
  \multicolumn{1}{c|}{\begin{tabular}[c]{@{}c@{}}English,\\ German,\\ Italian,\\ Latin,\\ Norwegian,\\ Russian,\\ Swedish\end{tabular}} \\ \hline
\multicolumn{1}{|c}{\citeauthor{ryzhova2021detection}} &
  time-oblivious &
  unsupervised &
  \begin{tabular}[c]{@{}c@{}}ELMo,\\ RuBERT\end{tabular} &
  \begin{tabular}[c]{@{}c@{}}monolingual,\\ multilingual\end{tabular} &
  \begin{tabular}[c]{@{}c@{}}pre-trained\\ trained\end{tabular} &
  - &
  - &
  - &
  APD &
  \multicolumn{1}{c|}{Russian} \\ \hline
\multicolumn{1}{|c}{\citeauthor{kutuzov2022contextualized}} &
  time-oblivious &
  unsupervised &
  \begin{tabular}[c]{@{}c@{}}BERT-base,\\ ELMo\end{tabular} &
  \begin{tabular}[c]{@{}c@{}}monolingual,\\ multilingual\end{tabular} &
  domain adaptation &
  last &
  - &
  - &
  \begin{tabular}[c]{@{}c@{}}APD,\\ PRT\end{tabular} &
  \multicolumn{1}{c|}{\begin{tabular}[c]{@{}c@{}}English,\\ German,\\ Latin,\\ Swedish\end{tabular}} \\ \hline
\multicolumn{1}{|c}{\citeauthor{rachinskiy2021zeroshot}} &
  time-oblivious &
  supervised &
  XLM-R-base &
  multilingual &
  \begin{tabular}[c]{@{}c@{}}fine-tuned,\\ pre-trained\end{tabular} &
  - &
  - &
  - &
  APD &
  \multicolumn{1}{c|}{Russian} \\ \hline
\multicolumn{1}{|c}{\citeauthor{rosin2022temporal}} &
  time-aware &
  unsupervised &
  BERT-base &
  monolingual &
  fine-tuned &
  - &
  - &
  - &
  CD &
  \multicolumn{1}{c|}{\begin{tabular}[c]{@{}c@{}}English,\\ Latin,\\ German\end{tabular}} \\ \hline
\end{tabular}%
}
\caption{Summary view of ensemble approaches. Missing information is denoted with a dash}
\label{tab:ensemble}
\end{table}

According to Table~\ref{tab:ensemble}, we note that all the ensemble approaches are time-oblivious with the exception of~\cite{pomsl2020circe} and~\cite{rosin2022temporal}. We also note that unsupervised learning modalities are adopted with the exception of~\cite{rachinskiy2021zeroshot}. As a further remark, most of the ensemble solutions exploit models trained over different languages.

Some ensemble approaches combine form-based and sense-based measures to improve the quality of results. On the one hand, form-based measures are exploited to better capture the dominant sense of the target word $w$. On the other hand, sense-based measures are exploited to represent all the meanings of $w$, including the minor ones. 
The combination of CD (see form-based approaches in Section~\ref{sec:form}) and JSD (see sense-based approaches in Section~\ref{sec:sense}) is proposed in~\cite{martinc2020discovery}. As a further ensemble experiment, the results of combining APD, HD, and JSD are discussed in~\cite{wang2020university}. The APD measure is also considered in~\cite{rachinskiy2021zeroshot}, where multiple shift scores are calculated by using different distance metrics (e.g., Manatthan distance, CD, euclidean distance) and these scores are exploited to train a regression model as an ensemble.  

Ensemble approaches based on two form-based measures are also proposed. For instance, in~\cite{giulianelli2022fire}, the final semantic shift $s_w$ is obtained by averaging APD and PRT scores. This is motivated by experimental results where sometimes APD outperforms PRT, while some other times PRT outperforms APD~\citep{kutuzov2020uio}. 

Some other ensemble approaches are based on the idea to combine static and contextualised embeddings. The intuition is that static embeddings can capture the dominant sense of the target word $w$, better than form-based, contextualised embeddings. In~\cite{pomsl2020circe,teodorescu2022ualberta}, the semantic shift $s_w$ is assessed by leveraging both static and contextualised embeddings. In particular, $s_w$ is determined by the linear combination of the scores obtained by two approaches: i) the APD measure over contextualised embeddings (see form-based approaches in Section~\ref{sec:form}); ii) the CD measure over static embeddings aligned according to the approach described in~\cite{hamilton2016diachronic}. Similarly, in~\cite{martinc2020discovery}, instead of directly using the APD measure, JSD is exploited over clusters of contextualised embeddings (see sense-based approaches in Section~\ref{sec:sense}). As a further difference, the scores obtained by static and contextualised approaches are combined by multiplication. The intuition is that, since the score distributions of the two approaches are unknown, multiplication prevents an approach from contributing more than the other one in the final score.

\app\ approaches can be also combined with grammatical profiles under the intuition that grammatical changes are slow and gradual, while lexical contexts can change very quickly~\cite{kutuzov2021grammatical,giulianelli2022fire}. Grammatical profile vectors  $gp_w^1$ and $gp_w^2$ are associated with the times $t_1$ and $t_2$, respectively, to represent morphological and syntactical features of the considered language in the time period. In~\cite{ryzhova2021detection}, the contextualised embeddings of the word $w$ occurrences are combined with the grammatical vectors. A linear regression model with regularisation is trained by using as features the cosine similarities over $\Phi_w^1$ and $\Phi_w^2$, and over the grammatical vectors $gp_w^1$ and $gp_w^2$. 

As a further ensemble approach, the combination of the time-aware techniques presented in~\cite{rosin2022temporal} and~\cite{rosin2021time} (see form-based approaches in Section~\ref{sec:form}) is proposed in order to better inject time into word embeddings.

\subsection{Discussion}\label{sec:discussion}
According to Section~\ref{sec:form},~\ref{sec:sense}, and~\ref{sec:ensemble}, we note that form-based approaches are more popular than sense-based ones. Most papers are characterised by time-oblivious approaches and only a few time-aware approaches have recently appeared (e.g.,~\cite{rosin2022temporal}). All approaches leverage unsupervised learning modalities with few exceptions (e.g.,~\cite{hu2019diachronic}). We argue that the motivation is due to the recent introduction of a reference evaluation framework for semantic shift assessment proposed at SemEval Shared Task 1, where participants were asked to adopt an unsupervised configuration~\citep{schlechtweg2020semeval}.

All papers are featured by contextualised word embeddings extracted from BERT-like models. Regardless of their version (i.e., tiny, small, base, large), BERT and XLM-R are the most frequently used models, and only a few experiments rely on ELMo and RoBERTa. As a matter of fact, the size of data needed to train or fine-tune an XLM-R model is several orders of magnitude greater than BERT. Moreover, even if less frequently employed than BERT, ELMo seems to be promising for \app\ and outperform BERT, while being much faster in training and inference~\citep{kutuzov2020uio}. As a further interesting remark, the use of static \textit{document} embeddings extracted from a Doc2Vec model has been proposed to provide pseudo-contextualised \textit{word} embeddings as an alternative to BERT~\citep{periti2022what}.

Monolingual and multilingual language models are both popular. The BERT models are the most frequently used monolingual models. XLM-R models are generally preferred to mBERT (i.e., multilingual BERT) models, since the former are trained on a larger amount of data and languages, thus the intuition is that they can better encode the language usages. Multilingual models are used both in multilingual settings, where corpora of different languages are considered (e.g.,~\cite{martinc2020leveraging}), and monolingual settings, where just corpora of one language are given (e.g., in~\cite{giulianelli2022fire}). In a monolingual setting, the use of a multilingual model is motivated by two reasons: i) a model pre-trained on a specific language is not available (e.g.,~\cite{kutuzov2020uio}), ii) multilingual models are employed to exploit their cross-lingual transferability property (e.g.,~\cite{rachinskiy2021zeroshot}).

About the type of training, most of the papers directly use pre-trained models or fine-tune them for domain adaptation. Only a few papers propose to exploit a specific fine-tuning (e.g.,~\cite{pomsl2020circe}) or to incrementally fine-tune a pre-trained model (e.g.,~\cite{kutuzov2020uio}). Experiments indicate that fine-tuning a pre-trained model for domain adaptation consistently boosts the quality of results when compared against pre-trained models (e.g.,~\cite{qiu2022histbert}). The impact of fine-tuning on performance is analysed in~\cite{martinc2020capturing}, where it is shown that optimal results are achieved by fine-tuning a pre-trained model for five epochs and that, after five epochs, performance decreases due to over-fitting. However, we argue that the fine-tuning effectiveness strictly depends on the size and domain of the considered corpora. In many papers, a different number of epochs is proposed with varying results (e.g.,~\cite{kutuzov2020uio}).

When a transformer-based model is used, contextualised word embeddings are typically extracted from the last one or the last four layers of the model. Experiments show that the semantic features of text are mainly encoded in the last four encoder layers of BERT~\citep{jawahar2019bert,devlin2019bert}. In some papers, contextualised embeddings are extracted by aggregating the output of the first and the last encoded layers. In this case, the idea is to combine \textit{surface} features (i.e., phrase-level information~\citep{jawahar2019bert}) encoded in the first layer with the semantic features from the last one. Only in~\cite{laicher2021explaining}, the standalone use of lower layers of BERT is proposed. Middle layers of BERT are usually excluded since they mainly encode syntactic features~\cite{jawahar2019bert}. When contextualised embeddings are extracted from more than one layer, they are generally aggregated by average or sum (e.g,~\cite{periti2022what}). As an alternative, the use of concatenation is proposed in~\cite{kanjirangat2020sst}. 

As a further note, when a BERT-like model is used, some words may be split into word pieces by a subword-based tokenisation algorithm~\citep{sennrich2016neural,wu2016google}. In this case, word piece representations are generally synthesised into a single word representation $e_{w,k}^j$ through averaging (e.g.,~\cite{martinc2020leveraging}), or concatenating (e.g.,~\cite{martinc2020discovery}). As alternative to avoid such problem, the pre-trained vocabulary associated with the model can be extended by adding some words of interest. Then, a fine-tuning step is performed in order to learn the weights associated with the added words (e.g.,~\cite{rosin2021time}).

Clustering operations are typically exploited in sense-based approaches to perform Word Sense Induction~\citep{lau2012word}. The only form-based approach that relies on clustering is presented in~\cite{beck2020diasense} (see Section~\ref{sec:form} for details). The clustering algorithms that are most frequently employed are K-Means and affinity propagation (AP). %Their application is related to different reasons. On one hand, K-Means requires to define the number of clusters in advance; this is questionable under the assumption that each cluster denotes a specific word meaning of a target word because the number of word meanings is not known beforehand and different words may have a different number of word meanings. On the other hand,
Further considered clustering algorithms are Gaussian Mixture Models (GMMs) (e.g.,~\cite{rother2020cmce}),
agglomerative clustering (AGG) (e.g.,~\cite{arefyev2020bos}),
DBSCAN (e.g.,~\cite{karnysheva2020tue}), HDBSCAN (e.g.,~\cite{rother2020cmce}), Balanced Iterative Reducing and Clustering using Hierarchies (BIRCH) (e.g.,~\cite{rother2020cmce}), A-Posteriori affinity Propagation (APP) (e.g.,~\cite{periti2022what}), and Incremental Affinity Propagation based on Nearest neighbor Assignment (IAPNA) (e.g.,~\cite{periti2022what}). Since K-Means, GMMs, and AGG require to define the number of clusters in advance, the use of a silhouette score is generally employed to determine the optimal number of clusters. As an alternative, the AP algorithm is employed to let emerge the number of clusters without prefixing it. DBSCAN is proposed due to its capability of reducing noise by specifying i) the minimum number of embeddings of each cluster, and ii) the maximum distance $\epsilon$ between two embeddings in a cluster. HDBSCAN is the hierarchical version of DBSCAN and it can manage clusters of different sizes. As a difference with DBSCAN, HDBSCAN can detect noise without the $\epsilon$ parameter. APP and IAPNA are incremental extensions of AP, and their use is proposed for semantic shift detection when more than one time interval is considered. In~\cite{rother2020cmce}, different clustering algorithms are compared and the experiments show that i) DBSCAN is very sensitive to scale since $\epsilon$ is prefixed, and ii) BIRCH tends to find a lot of small clusters that are marginal with respect to word meanings.

About the shift functions, a detailed presentation of possible alternatives has been provided in Sections~\ref{sec:form} and~\ref{sec:sense}. As a final remark, we note that CD and APD are frequently exploited in form-based approaches, while JSD is commonly employed in sense-based approaches. 

Finally, as for the language of considered corpora, most papers consider the shared benchmark datasets taken from competitive evaluation campaigns (e.g., LSCDiscovery~\citep{zamora2022lscdiscovery}). Common considered languages are English, German, Latin, and Swedish that appeared in 2020 at SemEval Task 1. Russian appeared in 2021 at RuShiftEval. Spanish appeared in 2022 at LSCDiscovery. The Italian language was introduced in 2020 at DIACRIta, even if it did not receive much attention. The approach described in~\cite{martinc2020leveraging} represents a novel attempt to consider a diachronic corpus containing texts of different languages, namely English and Slovenian.

\section{Comparison of approaches on performances}~\label{sec:comparison}

\begin{table}[!ht]
\centering
\resizebox{\textwidth}{!}{%
\begin{tabular}{cccccccccccccccc|}
\multicolumn{1}{c|}{\textbf{}} &
  \multicolumn{1}{c|}{\textbf{\begin{tabular}[c]{@{}c@{}}SemEval\\ Englsh\end{tabular}}} &
  \multicolumn{1}{c|}{\textbf{\begin{tabular}[c]{@{}c@{}}SemEval\\ German\end{tabular}}} &
  \multicolumn{1}{c|}{\textbf{\begin{tabular}[c]{@{}c@{}}SemEval\\ Latin\end{tabular}}} &
  \multicolumn{1}{c|}{\textbf{\begin{tabular}[c]{@{}c@{}}SemEval\\ Swedish\end{tabular}}} &
  \multicolumn{1}{c|}{\textbf{\begin{tabular}[c]{@{}c@{}}GEMS\\ English\end{tabular}}} &
  \multicolumn{1}{c|}{\textbf{\begin{tabular}[c]{@{}c@{}}LivFC\\ English\end{tabular}}} &
  \multicolumn{1}{c|}{\textbf{\begin{tabular}[c]{@{}c@{}}COHA\\ English\end{tabular}}} &
  \multicolumn{1}{c|}{\textbf{\begin{tabular}[c]{@{}c@{}}LSCD\\ Spanish\end{tabular}}} &
  \multicolumn{1}{c|}{\textbf{\begin{tabular}[c]{@{}c@{}}DURel\\ German\end{tabular}}} &
  \multicolumn{1}{c|}{\textbf{\begin{tabular}[c]{@{}c@{}}SURel\\ German\end{tabular}}} &
  \multicolumn{3}{c|}{\textbf{\begin{tabular}[c]{@{}c@{}}RSE\\ Russian\end{tabular}}} &
  \multicolumn{2}{c|}{\textbf{\begin{tabular}[c]{@{}c@{}}NOR\\ Norwegian\end{tabular}}} \\
\multicolumn{1}{c|}{\textbf{Ref.}} &
  \multicolumn{1}{c|}{$C_1 - C_2$} &
  \multicolumn{1}{c|}{$C_1 - C_2$} &
  \multicolumn{1}{c|}{$C_1 - C_2$} &
  \multicolumn{1}{c|}{$C_1 - C_2$} &
  \multicolumn{1}{c|}{$C_1 - C_2$} &
  \multicolumn{1}{c|}{$C_1 - C_2$} &
  \multicolumn{1}{c|}{$C_1 - C_2$} &
  \multicolumn{1}{c|}{$C_1 - C_2$} &
  \multicolumn{1}{c|}{$C_1 - C_2$} &
  \multicolumn{1}{c|}{$C_1 - C_2$} &
  $C_1 - C_2$ &
  $C_2 - C_3$ &
  \multicolumn{1}{c|}{$C_1 - C_3$} &
  $C_1 - C_2$ &
  $C_2 - C_3$ \\ \hline
\multicolumn{1}{|c}{\citeauthor{teodorescu2022ualberta}} &
  - &
  - &
  - &
  - &
  - &
  - &
  - &
  \begin{tabular}[c]{@{}c@{}}ensemble\\ APD\\ 0.573\end{tabular} &
  - &
  - &
  - &
  - &
  - &
  - &
  - \\ \hline
\multicolumn{1}{|c}{\citeauthor{zhou2020temporalteller}} &
  \begin{tabular}[c]{@{}c@{}}form-based\\ CD\\ $0.392$\end{tabular} &
  \begin{tabular}[c]{@{}c@{}}form-based\\ CD\\ $0.392$\end{tabular} &
  \begin{tabular}[c]{@{}c@{}}form-based\\ CD\\ $0.392$\end{tabular} &
  \begin{tabular}[c]{@{}c@{}}form-based\\ CD\\ $0.392$\end{tabular} &
  - &
  - &
  - &
  - &
  - &
  - &
  - &
  - &
  - &
  - &
  - \\ \hline
\multicolumn{1}{|c}{\citeauthor{montariol2021scalable}} &
  \begin{tabular}[c]{@{}c@{}}sense-based\\ AP + WD\\ 0.456\end{tabular} &
  \begin{tabular}[c]{@{}c@{}}sense-based\\ AP + JSD\\ 0.583\end{tabular} &
  \begin{tabular}[c]{@{}c@{}}form-based\\ CD\\ 0.496\end{tabular} &
  \begin{tabular}[c]{@{}c@{}}sense-based\\ K-Means + WD\\ 0.332\end{tabular} &
  \begin{tabular}[c]{@{}c@{}}\textbf{sense-based}\\ \textbf{AP + JSD}\\ \textbf{0.510}\end{tabular} &
  - &
  - &
  - &
  \begin{tabular}[c]{@{}c@{}}sense-based\\ AP + JSD\\ 0.712\end{tabular} &
  - &
  - &
  - &
  - &
  - &
  - \\ \hline
\multicolumn{1}{|c}{\citeauthor{periti2022what}} &
  \begin{tabular}[c]{@{}c@{}}sense-based\\ AP + JSD\\ 0.514*\end{tabular} &
  - &
  \begin{tabular}[c]{@{}c@{}}sense-based\\ APP + JSD\\ 0.512*\end{tabular} &
  - &
  - &
  - &
  - &
  - &
  - &
  - &
  - &
  - &
  - &
  - &
  - \\ \hline
\multicolumn{1}{|c}{\citeauthor{pomsl2020circe}} &
  \begin{tabular}[c]{@{}c@{}}ensemble\\ APD\\ 0.246\end{tabular} &
  \begin{tabular}[c]{@{}c@{}}ensemble\\ APD\\ 0.725\end{tabular} &
  \begin{tabular}[c]{@{}c@{}}ensemble\\ APD\\ 0.463\end{tabular} &
  \begin{tabular}[c]{@{}c@{}}ensemble\\ APD\\ 0.546\end{tabular} &
  - &
  - &
  - &
  - &
  \begin{tabular}[c]{@{}c@{}}\textbf{ensemble}\\ \textbf{APD}\\ \textbf{0.802}\end{tabular} &
  \begin{tabular}[c]{@{}c@{}}\textbf{ensemble}\\ \textbf{APD}\\ \textbf{0.723}\end{tabular} &
  - &
  - &
  - &
  - &
  - \\ \hline
\multicolumn{1}{|c}{\citeauthor{rachinskiy2021zeroshot}} &
  - &
  - &
  - &
  - &
  - &
  - &
  - &
  - &
  - &
  - &
  \begin{tabular}[c]{@{}c@{}}ensemble\\ APD\\ 0.781\end{tabular} &
  \begin{tabular}[c]{@{}c@{}}ensemble\\ APD\\ 0.803\end{tabular} &
  \begin{tabular}[c]{@{}c@{}}ensemble\\ APD\\ 0.822\end{tabular} &
  - &
  - \\ \hline
\multicolumn{1}{|c}{\citeauthor{rachinskiy2022glossreader}} &
  - &
  - &
  - &
  - &
  - &
  - &
  - &
  \begin{tabular}[c]{@{}c@{}}\textbf{sense}\\ \textbf{APDP}\\ \textbf{0.745}\end{tabular} &
  - &
  - &
  - &
  - &
  - &
  - &
  - \\ \hline
\multicolumn{1}{|c}{\citeauthor{rodina2020elmo}} &
  - &
  - &
  - &
  - &
  - &
  - &
  - &
  - &
  - &
  - &
  \begin{tabular}[c]{@{}c@{}}form-based\\ PRT\\ 0.557\end{tabular} &
  \begin{tabular}[c]{@{}c@{}}sense-based\\ AP + JSD\\ 0.406\end{tabular} &
  - &
  - &
  - \\ \hline
\multicolumn{1}{|c}{\citeauthor{rosin2021time}} &
  \begin{tabular}[c]{@{}c@{}}form-based\\ CD\\ 0.467\end{tabular} &
  - &
  \begin{tabular}[c]{@{}c@{}}form-based\\ CD\\ 0.512\end{tabular} &
  - &
  - &
  \begin{tabular}[c]{@{}c@{}}\textbf{form-based}\\ \textbf{TD}\\ \textbf{0.620}\end{tabular} &
  - &
  - &
  - &
  - &
  - &
  - &
  - &
  - &
  - \\ \hline
\multicolumn{1}{|c}{\citeauthor{rosin2022temporal}} &
  \begin{tabular}[c]{@{}c@{}}form-based\\ CD\\ 0.627\end{tabular} &
  \begin{tabular}[c]{@{}c@{}}form-based\\ CD\\ 0.763\end{tabular} &
  \begin{tabular}[c]{@{}c@{}}\textbf{form-based}\\ \textbf{CD}\\ \textbf{0.565}\end{tabular} &
  - &
  - &
  - &
  - &
  - &
  - &
  - &
  - &
  - &
  - &
  - &
  - \\ \hline
\multicolumn{1}{|c}{\citeauthor{rother2020cmce}} &
  \begin{tabular}[c]{@{}c@{}}sense-based\\ HDBSCAN\\ 0.512\end{tabular} &
  \begin{tabular}[c]{@{}c@{}}sense-based\\ GMMs\\ 0.605\end{tabular} &
  \begin{tabular}[c]{@{}c@{}}sense-based\\ GMMs\\ 0.321\end{tabular} &
  \begin{tabular}[c]{@{}c@{}}sense-based\\ HDBSCAN\\ 0.308\end{tabular} &
  - &
  - &
  - &
  - &
  - &
  - &
  - &
  - &
  - &
  - &
  - \\ \hline
\multicolumn{1}{|c}{\citeauthor{ryzhova2021detection}} &
  - &
  - &
  - &
  - &
  - &
  - &
  - &
  - &
  - &
  - &
  \begin{tabular}[c]{@{}c@{}}ensemble\\ regression\\ 0.480*\end{tabular} &
  \begin{tabular}[c]{@{}c@{}}ensemble\\ regression\\ 0.487*\end{tabular} &
  \begin{tabular}[c]{@{}c@{}}ensemble\\ regression\\ 0.560*\end{tabular} &
  - &
   \\ \hline
\multicolumn{1}{|c}{\citeauthor{kudisov2022bos}} &
  - &
  - &
  - &
  - &
  - &
  - &
  - &
  \begin{tabular}[c]{@{}c@{}}form-based\\ APD\\ 0.637\end{tabular} &
  - &
  - &
  - &
  - &
  - &
  - &
  - \\ \hline
\multicolumn{1}{|c}{\citeauthor{kutuzov2020distributional}} &
  \begin{tabular}[c]{@{}c@{}}form-based\\ APD\\ 0.605\end{tabular} &
  \begin{tabular}[c]{@{}c@{}}form-based\\ PRT\\ 0.740\end{tabular} &
  \begin{tabular}[c]{@{}c@{}}form-based\\ PRT\\ 0.561\end{tabular} &
  \begin{tabular}[c]{@{}c@{}}\textbf{form-based}\\ \textbf{APD}\\ \textbf{0.610}\end{tabular} &
  \begin{tabular}[c]{@{}c@{}}sense-based\\ AP + JSD\\ 0.456*\end{tabular} &
  - &
  - &
  - &
  - &
  - &
  - &
  - &
  - &
  - &
  - \\ \hline
\multicolumn{1}{|c}{\citeauthor{laicher2021explaining}} &
  \begin{tabular}[c]{@{}c@{}}form-based\\ APD\\ 0.571*\end{tabular} &
  \begin{tabular}[c]{@{}c@{}}form-based\\ CD\\ 0.755*\end{tabular} &
  - &
  \begin{tabular}[c]{@{}c@{}}form-based\\ APD\\ 0.602*\end{tabular} &
  - &
  - &
  - &
  - &
  - &
  - &
  - &
  - &
  - &
  - &
  - \\ \hline
\multicolumn{1}{|c}{\citeauthor{liu2021statistically}} &
  \begin{tabular}[c]{@{}c@{}}form-based\\ CD\\ 0.341\end{tabular} &
  \begin{tabular}[c]{@{}c@{}}form-based\\ CD\\ 0.512\end{tabular} &
  \begin{tabular}[c]{@{}c@{}}form-based\\ CD\\ 0.304\end{tabular} &
  \begin{tabular}[c]{@{}c@{}}form-based\\ CD\\ 0.304\end{tabular} &
  \begin{tabular}[c]{@{}c@{}}form-based\\ CD\\ 0.286\end{tabular} &
  \begin{tabular}[c]{@{}c@{}}form-based\\ CD\\ 0.561\end{tabular} &
  - &
  - &
  - &
  - &
  - &
  - &
  - &
  - &
  - \\ \hline
\multicolumn{1}{|c}{\citeauthor{martinc2020discovery}} &
  \begin{tabular}[c]{@{}c@{}}ensemble\\ AP + JSD\\ 0.361\end{tabular} &
  \begin{tabular}[c]{@{}c@{}}ensemble\\ AP + JSD\\ 0.642\end{tabular} &
  \begin{tabular}[c]{@{}c@{}}form-based\\ CD\\ 0.496\end{tabular} &
  \begin{tabular}[c]{@{}c@{}}ensemble\\ AP + JSD\\ 0.343\end{tabular} &
  - &
  - &
  - &
  - &
  - &
  - &
  - &
  - &
  - &
  - &
  - \\ \hline
\multicolumn{1}{|c}{\citeauthor{giulianelli2020analysing}} &
  - &
  - &
  - &
  - &
  \begin{tabular}[c]{@{}c@{}}form-based\\ APD\\ 0.285*\end{tabular} &
  - &
  - &
  - &
  - &
  - &
  - &
  - &
  - &
  - &
  - \\ \hline
\multicolumn{1}{|c}{\citeauthor{giulianelli2022fire}} &
  \begin{tabular}[c]{@{}c@{}}form-based\\ APD\\ 0.514\end{tabular} &
  \begin{tabular}[c]{@{}c@{}}ensemble\\ PRT\\ 0.354\end{tabular} &
  \begin{tabular}[c]{@{}c@{}}ensemble\\ PRT\\ 0.572\end{tabular} &
  \begin{tabular}[c]{@{}c@{}}ensemble\\ APD\\ 0.397\end{tabular} &
  - &
  - &
  - &
  - &
  - &
  - &
  \begin{tabular}[c]{@{}c@{}}ensemble\\ APD + PRT\\ 0.376\end{tabular} &
  \begin{tabular}[c]{@{}c@{}}form-based\\ APD\\ 0.480\end{tabular} &
  \begin{tabular}[c]{@{}c@{}}form-based\\ APD\\ 0.457\end{tabular} &
  \begin{tabular}[c]{@{}c@{}}\textbf{ensemble}\\ \textbf{APD + PRT}\\ \textbf{0.394}\end{tabular} &
  \begin{tabular}[c]{@{}c@{}}\textbf{ensemble}\\ \textbf{APD}\\ \textbf{0.503}\end{tabular} \\ \hline
\multicolumn{1}{|c}{\citeauthor{hu2019diachronic}} &
  - &
  - &
  - &
  - &
  - &
  - &
  \begin{tabular}[c]{@{}c@{}}\textbf{sense-based}\\ \textbf{MNS}\\ \textbf{0.428}*\end{tabular} &
  - &
  - &
  - &
  - &
  - &
  - &
  - &
  - \\ \hline
\multicolumn{1}{|c}{\citeauthor{kanjirangat2020sst}} &
  \begin{tabular}[c]{@{}c@{}}sense-based\\ K-Means + JSD\\ 0.028*\end{tabular} &
  \begin{tabular}[c]{@{}c@{}}sense-based\\ K-Means + JSD\\ 0.173*\end{tabular} &
  \begin{tabular}[c]{@{}c@{}}sense-based\\ K-Means + JSD\\ 0.253*\end{tabular} &
  \begin{tabular}[c]{@{}c@{}}sense-based\\ K-Means + CSC\\ 0.321*\end{tabular} &
  - &
  - &
  - &
  - &
  - &
  - &
  - &
  - &
  - &
  - &
  - \\ \hline
\multicolumn{1}{|c}{\citeauthor{karnysheva2020tue}} &
  \begin{tabular}[c]{@{}c@{}}sense-based\\ K-Means + JSD\\ -0.155*\end{tabular} &
  \begin{tabular}[c]{@{}c@{}}sense-based\\ DBSCAN + JSD\\ 0.388*\end{tabular} &
  \begin{tabular}[c]{@{}c@{}}sense-based\\ DBSCAN + JSD\\ 0.177*\end{tabular} &
  \begin{tabular}[c]{@{}c@{}}sense-based\\ K-Means + JSD\\ -0.062*\end{tabular} &
  - &
  - &
  - &
  - &
  - &
  - &
  - &
  - &
  - &
  - &
  - \\ \hline
\multicolumn{1}{|c}{\citeauthor{kashleva2022hse}} &
  - &
  - &
  - &
  - &
  - &
  - &
  - &
  \begin{tabular}[c]{@{}c@{}}sense-based\\ APDP\\ 0.553\end{tabular} &
  - &
  - &
  - &
  - &
  - &
  - &
  - \\ \hline
\multicolumn{1}{|c}{\citeauthor{keidar2022slangvolution}} &
  \begin{tabular}[c]{@{}c@{}}form-based\\ APD\\ 0.489\end{tabular} &
  - &
  - &
  - &
  - &
  - &
  - &
  - &
  - &
  - &
  - &
  - &
  - &
  - &
  - \\ \hline
\multicolumn{1}{|c}{\citeauthor{arefyev2021deep}} &
  - &
  - &
  - &
  - &
  - &
  - &
  - &
  - &
  - &
  - &
  \begin{tabular}[c]{@{}c@{}}\textbf{form-based}\\ \textbf{APD}\\ \textbf{0.825}\end{tabular} &
  \begin{tabular}[c]{@{}c@{}}form-based\\ APD\\ 0.821\end{tabular} &
  \begin{tabular}[c]{@{}c@{}}form-based\\ APD\\ 0.823\end{tabular} &
  - &
  - \\ \hline
\multicolumn{1}{|c}{\citeauthor{arefyev2020bos}} &
  \begin{tabular}[c]{@{}c@{}}sense-based\\ AGG + CD\\ 0.299\end{tabular} &
  \begin{tabular}[c]{@{}c@{}}sense-based\\ AGG + CD\\ 0.094\end{tabular} &
  \begin{tabular}[c]{@{}c@{}}sense-based\\ AGG + CD\\ -0.134\end{tabular} &
  \begin{tabular}[c]{@{}c@{}}sense-based\\ AGG + CD\\ 0.274\end{tabular} &
  - &
  - &
  - &
  - &
  - &
  - &
  - &
  - &
  - &
  - &
  - \\ \hline
\multicolumn{1}{|c}{\citeauthor{beck2020diasense}} &
  \begin{tabular}[c]{@{}c@{}}form-based\\ CD\\ 0.293*\end{tabular} &
  \begin{tabular}[c]{@{}c@{}}form-based\\ CD\\ 0.414*\end{tabular} &
  \begin{tabular}[c]{@{}c@{}}form-based\\ CD\\ 0.343*\end{tabular} &
  \begin{tabular}[c]{@{}c@{}}form-based\\ CD\\ 0.300*\end{tabular} &
  - &
  - &
  - &
  - &
  - &
  - &
  - &
  - &
  - &
  - &
  - \\ \hline
\multicolumn{1}{|c}{\citeauthor{cuba2020sensecluster}} &
  \begin{tabular}[c]{@{}c@{}}form-based\\ CD\\ 0.209*\end{tabular} &
  \begin{tabular}[c]{@{}c@{}}form-based\\ CD\\ 0.656*\end{tabular} &
  \begin{tabular}[c]{@{}c@{}}form-based\\ CD\\ 0.399*\end{tabular} &
  \begin{tabular}[c]{@{}c@{}}form-based\\ CD\\ 0.234*\end{tabular} &
  - &
  - &
  - &
  - &
  - &
  - &
  - &
  - &
  - &
  - &
  - \\ \hline
  \multicolumn{1}{|c}{\citeauthor{kutuzov2022contextualized}} &
  \begin{tabular}[c]{@{}c@{}}form-based\\ APD\\ 0.605\end{tabular} &
  \begin{tabular}[c]{@{}c@{}}form-based\\ PRT\\ 0.740\end{tabular} &
  \begin{tabular}[c]{@{}c@{}}form-based\\ PRT\\ 0.561\end{tabular} &
  \begin{tabular}[c]{@{}c@{}}form-based\\ APD\\ 0.569\end{tabular} &
  \begin{tabular}[c]{@{}c@{}}form-based\\ APD\\ 0.394\end{tabular} &
  - &
  - &
  - &
  - &
  - &
  - &
  - &
  - &
  - &
  - \\ \hline
  \multicolumn{1}{|c}{\citeauthor{periti2023studying}} &
  \begin{tabular}[c]{@{}c@{}}sense-based\\ APP + APDP\\ 0.651*\end{tabular} &
  \begin{tabular}[c]{@{}c@{}}sense-based\\ APP + APDP\\ 0.527*\end{tabular} &
  \begin{tabular}[c]{@{}c@{}}sense-based\\ APP + JSD\\ 0.433*\end{tabular} &
  \begin{tabular}[c]{@{}c@{}}sense-based\\ APP + APDP\\ 0.499*\end{tabular} &
  - &
  - &
  - &
  \begin{tabular}[c]{@{}c@{}}sense-based\\ APP + APDP\\ 0.544*\end{tabular} & 
  - &
  - &
  \begin{tabular}[c]{@{}c@{}}sense-based\\ APP + APDP\\ 0.273*\end{tabular} &
  \begin{tabular}[c]{@{}c@{}}sense-based\\ APP + APDP\\ 0.393*\end{tabular} &
  \begin{tabular}[c]{@{}c@{}}sense-based\\ APP + APDP\\ 0.407*\end{tabular} &
  - &
  - \\ \hline
  \multicolumn{1}{|c}{\citeauthor{cassotti2023xl}} &
  \begin{tabular}[c]{@{}c@{}}\textbf{form-based}\\ \textbf{APD}\\ \textbf{0.757}\end{tabular} &
  \begin{tabular}[c]{@{}c@{}}\textbf{form-based}\\ \textbf{APD}\\ \textbf{0.877}\end{tabular} &
  \begin{tabular}[c]{@{}c@{}}form-based\\ APD\\ -0.056\end{tabular} &
  \begin{tabular}[c]{@{}c@{}}\textbf{form-based}\\ \textbf{APD}\\ \textbf{0.754}\end{tabular} &
  - &
  - &
  - &
  - &
  - &
  - &
  \begin{tabular}[c]{@{}c@{}}form-based\\ APD\\ 0.799*\end{tabular} &
  \begin{tabular}[c]{@{}c@{}}\textbf{form-based}\\ \textbf{APD}\\  \textbf{0.833}\end{tabular} &
  \begin{tabular}[c]{@{}c@{}}\textbf{form-based}\\ \textbf{APD}\\  \textbf{0.842}\end{tabular} &
  - &
  - \\ \hline
\end{tabular}%
}
\caption{The Spearman's correlation score of \app\ approaches in selected experiments over corpora of different languages. For each corpus, the top performance is reported in bold. Asterisks denote experiments based on a pre-trained model}
\label{tab:comp}
\end{table}

In this section, we propose a comparison of the \app\ approaches based on their performance obtained by considering the evaluation framework  adopted in SSD tasks of shared competitions. The framework is based on a reference benchmark which contains a diachronic textual corpus in a given language. The framework is also characterised by a test-set of target words, where each word is associated with a continuous shift score (i.e., {\em gold score}) calculated on the basis of manual annotation. Different metrics are also defined in the framework to evaluate the performance of the approaches according to the kind of assessment question that the task aims to address, namely {\em Grade/Binary Change}, {\em Sense Gain/Loss} (see Section~\ref{sec:problem}). 

In Table~\ref{tab:comp}, we compare the \app\ approaches by considering the experiments on {\em Grade Change Detection} task performed and reported in the corresponding literature papers. In such a kind of task, the Spearman's correlation score is typically employed for assessing the performance of a given experiment by measuring the correlation between the predicted shift scores and the gold scores\footnote{In~\cite{montariol2021scalable}, as an alternative to the Spearman's correlation score, the {\em Discount Cumulative Gain} is proposed~\cite{montariol2021scalable}. However, most papers still use Spearman's, since it is currently employed in competitive shared tasks.}. The Spearman's correlation evaluates the monotonic relationship between the rank-order of the predicted scores and the gold ones. When multiple experiments are discussed in a paper, in Table~\ref{tab:comp}, we report the best Spearman's correlation score obtained.

In the comparison, twelve diachronic corpora are exploited. In particular, we consider: i) the four SemEval datasets for English (SemEval English), German (SemEval German), Latin (SemEval Latin), and Swedish (SemEval Swedish)~\citep{schlechtweg2020semeval}; ii) the English dataset proposed in~\cite{gulordava2011distributional} (GEMS English); iii) the English LiverpoolFC dataset proposed in~\cite{deltredicietal2019short} (LivFC English); iv) the COHA English dataset (COHA English); v) the LSCDiscovery dataset for Spanish (LSCD Spanish)~\citep{zamora2022lscdiscovery}; vi) the DURel dataset for German (DURel German)~\citep{schlechtweg2018diachronic}; vii) the SURel dataset for German (SURel German)~\citep{hatty2019surel}; viii) the RuShiftEval dataset for Russian (RSE Russian)~\citep{kutuzov2021rushifteval}; and ix) the NorDiaChange dataset for Norwegian (NOR Norwegian)~\citep{kutuzov2022nordiachange}. In Table~\ref{tab:comp}, for each corpus, we highlight when a single time interval $C_1 - C_2$ or two consecutive time intervals $C_1 - C_2$ and $C_2 - C_3$ are considered, respectively. As a further remark, we note that the RSE Russian corpus is the only case where a test-set for the time interval $C_1 - C_3$ as a whole is provided.

For the sake of readability, the performance according to the Spearman's correlation scores shown in Table~\ref{tab:comp} are labeled with the semantic shift function of the considered \app\ approach and the corresponding framing with respect to form-based, sense-based, and ensemble-based categories (see Section~\ref{sec:approaches}).

As a general remark, we cannot find an approach outperforming all the others on all the considered corpora. This can suggest that an approach is language-dependant, namely it works well on one language and it is not appropriate for others. By relying on the experiments presented in~\cite{kutuzov2020uio}, we claim that the approaches are not language-dependant and the performance of an approach is influenced by the employed assessment measure in relation with the distribution of the gold scores in the considered test-set. The experiments in~\cite{kutuzov2020uio} show that when the distribution of the gold scores is skewed, namely some words are highly shifted and some others are barely shifted, the APD measure achieves better performance on Spearman's correlation than the PRT measure. On the contrary, when the distribution of the gold scores is almost uniform, namely most of the words are similarly shifted, the PRT measure achieves better performance than the APD measure.

As a further remark, we note that the approaches characterised by fine-tuning achieve greater performance. This is also confirmed in the experiments of~\cite{martinc2020capturing} where fine-tuning a model boosts the performance when the model is not affected by under or over-fitting.

On average, form-based approaches outperform sense-based approaches in Grade Change Detection tasks. We argue that such a result is motivated by the structure of the test-sets, where just one semantic shift score is provided for each target word. Form-based approaches benefit from this structure since they work on measuring the shift over one general word property (i.e., the dominant sense, or the degree of polysemy). 
On the opposite, sense-based approaches are disadvantaged by this structure since they work on measuring the shift over multiple word meanings and they need to produce a single, comprehensive shift value that summarises all the single-meaning shifts for the comparison against the gold score. As a result, capturing some (minor) meanings can negatively affect the comprehensive shift value, and to address this issue, small clusters are usually considered as possible noise and filtered out~\citep{martinc2020discovery}.

Table~\ref{tab:comp} shows that form-based approaches based on APD, CD, or PRT measures tend to obtain higher performance than sense- and ensemble-based approaches. GEMS English, COHA English, and LSCD Spanish are the only benchmarks where sense-based approaches outperform form-based ones. This can be motivated by the small number of experiments performed. Indeed, for COHA English experiments with form-based approaches have not been tested~\citep{hu2019diachronic}, while only a few experiments and a limited number of configurations with form-based approaches have been tested on GEMS English. For LSCD Spanish, the top performance is 0.745 and the corresponding approach leverages the APDP measure, which is an extension of APD characterised by the use of an average-of-average operation. This result is in line with the intuition presented in~\cite{periti2022what}, where the use of averaging on top of clustering contributes to reduce the noise in the contextualised embeddings of the target word.

We also note that ensemble approaches are on average characterised by high performance. In particular, top performances are provided by ensemble approaches on DURel German (0.802), SURel German (0.723), and NOR Norwegian (0.394 and 0.503). It is interesting to observe that the performance on DURel and SURel German are obtained through an approach combining static and contextualised word embeddings, thus highlighting that such a kind of combination can be effective. For NOR Norwegian in the time interval $C_1 - C_2$, the best approach exploits both APD and PRT; this is a further confirmation that APD and PRT are top-performing measures in semantic shift detection. For the subsequent time interval $C_2 - C_3$, the best result on NOR Norwegian is obtained with a combination of APD with grammatical profiles. This is a confirmation of the intuition presented in~\cite{giulianelli2022fire}, which suggests that ensembling grammatical profiles with contextualised embeddings can enhance performance by incorporating morphological and syntactic features not fully captured by contextualised models.

For SemEval English, SemEval German, and SemEval Latin, the top performance are 0.627, 0.763, and 0.565, respectively, and they are obtained by the time-aware approach proposed in~\cite{rosin2022temporal}. Also for LivFC English (0.620), the top performance is obtained by leveraging a time-aware approach~\citep{rosin2021time}. We argue that extra-linguistic information (e.g., time information) can have a positive impact on performance. The injection of extra-linguistic information can contribute to increase the performance also when small-size language models are employed, since they are less affected by noise than larger models. As a confirmation, in contrast to the widespread belief that the larger the models the higher the performance, the best result for SemEval English is obtained by exploiting contextualised embeddings extracted from a BERT-tiny model~\citep{turc2019well,rosin2022temporal}. This is also true for SemEval Swedish (0.610), where the top performance is obtained by calculating the APD measure over contextualised embeddings extracted from an ELMo  model~\citep{kutuzov2020distributional}, which is far smaller than BERT-like models.

Finally, we note that the use of supervised learning modalities contributes to achieve high performance. As an example, the top performances for RSE Russian are 0.825 on $C_1-C_2$, 0.821 on $C_2-C_3$, and 0.823 on $C_1-C_3$ and they are obtained by a form-based, supervised approach~\citep{arefyev2021deep}.

\section{Scalability, interpretability, and robustness issues}\label{sec:limitations}
In this section, we analyse the \app\ approaches by considering possible scalability, interpretability, and reliability issues.

%%%%%%%%%%%%%%
\subsection{Scalability issues}\label{sub-sec:scal-issue}
In the \app\ approaches, any occurrence of the target word considered for shift assessment is represented by a specific embedding. As a basic implementation, all the contextualised embeddings are stored in memory for processing. The higher the number of occurrences of a target word, the higher the number of embeddings to manage. As a result, when the size of the diachronic corpus grows, possible issues arise both in terms of memory and computation time. Similar issues occur when multiple target words are considered for shift assessment. In this case, a possible workaround for addressing the memory issue is to process one target word at a time. However, in this way, the memory issue {\em shifts} to a computation time issue. For feasibility convenience, most experiments work on a small set of target words. This kind of limitations inhibits the possibility to address tasks like the detection of the most changed word in a corpus. The need to work on solutions capable of dealing with such a kind of scalability issues has recently been promoted in LSCDiscovery, where participants were asked to assess the semantic shift on all the words of the dictionary~\citep{zamora2022lscdiscovery}. 

Some possible solutions to the scalability issues have been proposed in literature. For instance, approaches based on measures that enforce aggregation by averaging (e.g., CD, PRT) are time-scalable, since only the prototypes are considered for shift assessment instead of the whole set of embeddings. Also approaches based on APD or JSD measures can be adjusted to become time-scalable. In particular, the number of embeddings to store and process can be reduced by random sampling the occurrences of the target word $w$. This means that i) a smaller number of similarity scores needs to be calculated with APD (e.g.,~\cite{ryzhova2021detection}), and ii) JSD works on top of clustering algorithms that converge faster (e.g.,~\cite{rodina2020elmo}). As an alternative to random sampling, an online {\em aggregation by summing} method is proposed in~\cite{montariol2021scalable}, where a pre-fixed number of contextualised embeddings $n$ is stored in memory. An embedding $e_w$ is stored when the number of embeddings in memory is less than $n$ and $e_w$ is strongly dissimilar from all the other embeddings previously stored. If $e_w$ is not stored, it is aggregated to the most similar embedding stored in memory through sum.

The dimensionality reduction of the embeddings is proposed as a further alternative to enforce scalability. For example, in~\cite{rother2020cmce}, the embedding dimensionality is reduced to 10 (from 768) by combining an autoencoder with the UMAP (Uniform Manifold Approximation and Projection) algorithm~\citep{mcinnes2018umap}. In~\cite{keidar2022slangvolution}, UMAP and PCA are used to project contextualised embedding into $h \in \{2, 5, 10, 20, 50, 100\}$ dimensions. With respect to this solution, we argue that, although it can improve the memory scalability, time scalability is negatively affected since  dimensionality reduction takes time. However, in~\cite{rother2020cmce}, it is shown that the dimensionality reduction can still contribute to time scalability when the goal is to test and compare the effectiveness of different clustering algorithms and the reduced embeddings are saved and re-used. As a further option, the use of small language models, such as TinyBert or ELMo, is gaining more and more attention since the dimension of the generated embeddings is far lower (e.g.,~\cite{rosin2022temporal}). 

Scalability issues can also arise when the shift needs to be assessed on a corpus $C = \bigcup_i^n C_i$ defined over more than one time interval ($n > 2$). Typically, \app\ approaches calculate the shift score $s_w$ over each pair of time intervals $(t_{i},t_{i+1})$ by iteratively re-applying the same assessment workflow. As a difference, an incremental approach based on a clustering algorithm called {\em A Posteriori affinity Propagation} (APP) is proposed in~\cite{periti2022what} to speed up the aggregation stage. In each time interval, clustering is incrementally executed by considering the prototypes of the previous time period (i.e., aggregation by averaging) and the incoming embeddings of the current time period.

%%%%%%%%%%%%%%
\subsection{Interpretability issues}
Interpretability issues arise when it is not possible to determine which meaning(s) have changed among all the meanings of a target word, namely the meaning(s) that mainly caused the shift score assessed by a considered approach. Definitely, form-based approaches are affected by such a kind of issues, since they model the shift as the change in the dominant sense or in the degree of polysemy of a word, without considering the possible multiple meanings. On the opposite, sense-based approaches aim at providing an interpretation of the word change, since they attempt to model the shift by considering the multiple word senses. However, interpretability issues can arise also when sense-based approaches are employed due to three main motivations.

\textit{Word meaning representation.}
Sense-based approaches mostly rely on clustering techniques to represent word meanings. The K-Means and the AP clustering algorithms are usually employed to this end. K-Means requires that the number of target clusters is prefixed, and this can be inappropriate to effectively represent the meanings of a target word that are not known beforehand. AP lets the number of target clusters emerge, but experimental results show that the association of a cluster with a word meaning can be imprecise. We argue that this can be due to the distributional nature of contextualised models that tends to capture changes in contextual variance (i.e., word usages) rather than changes in lexicographic senses (i.e., word meanings)~\citep{kutuzov2022contextualized}. As an example, sometimes AP produces more than 100 clusters, which is rather unrealistic if we assume that a cluster represents a word meaning~\citep{periti2022what}. As a matter of fact, a word may completely change its context without changing its meaning~\citep{martinc2020capturing}. 
%Since  the  resulting  semantic  changescore is a derivative of the arrays of token embeddings,one  observes  strong  bursts  which  manifest  changesin contextual variance of a word, not a semantic shift in  the  lexicographic  meaning  of  this  term.    This  isprobably not what a historical linguist expects to see,although it can depend on the particular study and theworking definition of ‘semantic shi’~\cite{kutuzov2022contextualized}.

\textit{Word meaning description.}
Each cluster obtained during the aggregation stage of a sense-based approach needs to be associated with a description that denotes the corresponding word meaning. This can be done by human experts on the basis of the cluster contents. However, this is time-consuming, given that a cluster can consist of several hundreds/thousands of elements. As an alternative, clustering analysis techniques have been proposed to label clusters by summarising their contents.
As a possible option, a cluster description can be extracted from the content by considering the top featuring keywords based on lexical occurrences (e.g., Tf-Idf)~\citep{kellert2022using,montariol2021scalable}. In~\cite{giulianelli2020analysing}, the sense-prototype of a cluster is proposed as a cluster exemplar and the corpus sentences that are closest to the prototype are adopted as cluster/meaning description. However, when a cluster contains outliers, these sentences could not provide an effective description. 

\textit{Word meaning evolution.}
When a corpus $C = \bigcup_i^n$ defined over more than one time interval is considered, the clusters defined at a time step $t_i$ need to be linked to the clusters of the previous time-step $t_{i-1}$ to trace the evolution of the corresponding meaning over time (i.e., cluster/meaning history). Since the clustering executions at each time-step are independent, the capability of recognising corresponding clusters/meanings at different time-steps can be challenging. As a possible solution, alignment techniques can be employed to link similar word meanings in different, consecutive time periods~\citep{kanjirangat2020sst,montariol2021scalable}. As a further option, evolutionary clustering algorithms can be exploited without requiring any alignment mechanism across time periods~\citep{periti2022what}.

%%%%%%%%%%%%%%
\subsection{Robustness issues}
Robustness issues arise when the assessment score is not reliable due to data imbalance, model stability, and model bias.

\textit{Data imbalance.}
The diachronic corpus $\mathcal{C}$ must equally reflect the presence of the target word $w$ in both the time steps $t_1$ and $t_2$. This means that the frequency of $w$ must not strongly change in the considered time period. However, in common scenarios, more documents are available for the most recent time step $t_2$ and \textit{``it may not be possible to achieve balance in the sense expected from a modern corpus''}~\citep{tahmasebi2018survey}. As a consequence, the frequency of $w$ can be strongly higher in $t_2$ than in $t_1$ and the embeddings $\Phi_w$ can produce a distorted representation of the target word when the model is trained/fine-tuned~\citep{zhou2021frequency,wendlandt2018factors}.
As a further remark, data imbalance issues can occur when some word meanings are more frequent than others. For instance, the dominant sense is usually more represented than other senses in the corpus $\mathcal{C}$. As a result, when a sense-based approach is adopted, the embedding distributions $p_w^1$, $p_w^2$ can be skewed, meaning that a larger number of embeddings is associated with the dominant sense rather than with the other minor senses. In sense-based approaches, the word meanings are represented by clusters, and \textit{the number of clusters consistently reflects word frequency}~\citep{kutuzov2020distributional}. When a meaning is associated with a few embeddings/clusters, its contribution to the overall assessment score is marginally leading to an inflated or underestimated assessment score. In this respect, a qualitative analysis of ``potentially erroneous'' outputs of \app\ approaches is presented in~\cite{kutuzov2022contextualized}. Some examples of potentially erroneous assessment scores occur when i) a \textit{word with strongly context-dependent meanings} is considered, whose embeddings are mutually different; ii) a \textit{word is frequently used in a very specific context} in only one time step $t_1$ or $t_2$; iii) a \textit{word is affected by a syntactic change}, not a semantic one. In~\cite{liu2021statistically}, a solution is proposed to reduce the false discovery rate and to improve the precision of the shift assessment by leveraging permutation-based statistical test and term-frequency thresholding.

\textit{Model stability.}
Contextualised pre-trained models are usually trained on modern text sources. For example, the original English BERT model is pre-trained on Wikipedia and BooksCorpus~\citep{zhu2015aligning}. As a result, pre-trained models are prone to represent words from a modern perspective, and thus they tend to ignore the temporal information of a considered corpus. This way, when historical corpora are considered, the possible obsolete word usages cannot be properly represented. This problem has been investigated in the literature by comparing the performance of pre-trained against fine-tuned models~\citep{kutuzov2020uio,qiu2022histbert}. In line with the considerations of Section~\ref{sec:discussion}, the results show that fine-tuning the model on the whole diachronic corpus improves the quality of word representations for historical texts. Since fine-tuning the model can be expensive in terms of time and computational resources, a measure for estimating the model effectiveness for historical sources is presented in~\cite{ishihara2022semantic}. In particular, in~\cite{ishihara2022semantic}, this measure is used to decide whether a model should be re-trained or fine-tuned.

\textit{Model bias}
The contextualised embeddings can possibly be affected by biases on the encoded information. For instance, a possible bias can arise from orthographic information, such as the word form and the position of a word in a sentence, since they influence the output of the top BERT layers~\citep{laicher2021explaining}. Text pre-processing techniques are proposed as a solution to reduce the influence of orthography in the embeddings, thus increasing the robustness of encoded semantic information. To this end, lower-casing the corpus text is a commonly-employed solution. However, \textit{the lower-casing of words often conflates parts of speech}, thus another possible bias can raise. For example, the proper noun \verb|Apple| and the common noun \verb|apple| become identical after lower-casing~\citep{hengchen2021challenges}. The possible bias introduced by Named Entities and proper nouns is investigated in~\cite{laicher2021explaining,martinc2020discovery}. In~\cite{qiu2022histbert}, text normalisation techniques are proposed based on the removal of accent markers. In some languages, such a kind of normalisation can introduce a bias since different words can be conflated. For example, \verb|papà| (e.g., the Italian word for \verb|dad|) and \verb|papa| (e.g., the Italian word for \verb|pope|) cannot be distinguished after the accent removal. %Due to the above considerations, the choice of applying text pre-processing techniques before embedding must be carefully considered.
Further text pre-processing techniques can be employed to reduce the possible bias due to orthographic information. In~\cite{schlechtweg2020semeval}, lemmatisation and punctuation removal are proposed. Experimental results on lemmatisation for reducing the model bias on BERT embeddings are presented in~\cite{laicher2021explaining}. Further experiments show that lemmatising the target word alone is more beneficial than lemmatising the whole corpus~\citep{laicher2021explaining}. Filtering out unimportant words, such as stop words and low-frequency words, can be also beneficial~\citep{zhou2020temporalteller}. As an alternative solution to reduce word-form biases, the embedding of a word occurrence can be computed by averaging its original embedding and the embeddings of its nearest words in the input sentence~\citep{zhou2020temporalteller}. 

When aggregation by clustering is enforced, the possible word-form biases can affect the clustering result~\citep{laicher2021explaining}. As a solution, clustering refinement techniques have been proposed. As an option, the removal of the clusters containing only one or two instances is adopted, since they are not considered significant~\citep{martinc2020discovery}. As a further option, in~\cite{martinc2020capturing}, clusters with less than two members are considered as weak clusters and they are merged with the closest strong cluster, i.e. cluster with more than two members. In~\cite{periti2022what}, clusters containing less than 5 percent of the whole set of embeddings are assumed to be poorly informative and are thus dropped. However, we argue that the use of clustering refinement techniques must be carefully considered since also small clusters can be important when the corpus is unbalanced in the number of meanings of a word.

\section{Challenges and concluding remarks}\label{sec:conclusion}
In this survey, we analysed the \app\ task by providing a formal definition of the problem and a reference classification framework based on meaning representation, time awareness, and learning modality dimensions. The literature approaches are surveyed according to the given framework by considering the assessment function, the language model, the achieved performance, and the possible scalability/interpretability/ro\-bust\-ness issues. 
% Although the  community interested in \app\ is growing, the use of contextualised embeddings for detecting semantic shifts is far from being consolidated. 

In~\cite{hengchen2021challenges}, an overview of open challenges about computational SSD is presented. In the following, we extend such an overview by focusing on those challenges that are specific to \app\ in relation with the issues discussed in Section~\ref{sec:limitations}.\\

%Although form-based approaches proved to outperform sense-based approaches in the Grade Change Detection task, they cannot be used to provide qualitative interpretations of which meaning change, as is the case for approaches based on static embeddings. In this regard and in light of the issues described in Section~\ref{sec:limitations}, since the use of static embeddings has been widely tested and its results are slightly lower than those of contextualised embeddings, a reasonable question can arise: is it really advantageous to use contextualised embeddings for semantic shift detection? We argue that their use is crucial for three reasons: i) the training of static embedding models is sub-optimal when a small-size dataset is considered. In this case, \app\ approaches based on a pre-trained model consistently outperform the use of static embeddings for semantic shift detection; ii) fine-tuning a contextualised pre-trained model definitely boosts the performance with respect to the results obtained through the use of static embeddings; finally, iii) sense-based approaches enable the interpretation of the detected semantic shift.

\textbf{Scalability.} 
The trend in \app\ is to adopt increasingly larger models with the idea that they better represent language features. As a consequence, scalability issues arise and they are being addressed as discussed in Section~\ref{sub-sec:scal-issue}. However, contrary to this trend, we argue that the use of small-size models, such as those introduced in~\cite{rosin2022temporal,rosin2021time}, needs to be further explored since they are competitive in terms of performance.\\

\textbf{Word meaning representation.} 
In Section~\ref{sec:comparison}, we show that form-based approaches outperform sense-based approaches in the Grade Change Detection assessment. However, we argue that sense-based approaches are promising since they focus on encoding word senses and they can enrich the mere degree of semantic shift of a word $w$ with the information about the specific meaning of $w$ that changed. In this direction, the SSD should be considered as a temporal/diachronic extension of other problems such as Word Sense Induction~\citep{alsulaimani2020evaluation}, Word Meaning Disambiguation~\citep{godbole2022temporal}, and Word-in-Context~\citep{loureiro2022tempowic}. 
%We note that \app\ should balance the role of linguistics and computer science since \app is far more computational than linguistic and no linguistics theory is generally considered. 
So far, word senses have been represented through aggregation by clustering under the idea that each cluster represents a specific word meaning. However, according to the interpretability issues of Section~\ref{sec:limitations}, clustering techniques are often affected by noise and they are typically capable of representing word usages rather than word meanings. Thus, further investigations are required to represent lexicographic meanings in a more faithful way. The possible integration of the linguistic theory should be also considered for sense modeling. For example, the linguistics theory could contribute to tailor the behavior of \app\ approaches by focusing on a specific type of word sense to consider (e.g., standard word meaning, topic use, pragmatics, connotation)~\citep{hengchen2021challenges}. As a further example, the linguistics theory can clarify the specific word meaning captured by a given cluster representation.\\

\textbf{Word meaning description.} 
According to Section~\ref{sec:limitations}, current solutions to meaning description are focused on determining a representative label taken from the cluster contents (e.g., Tf-Idf, sentence(s) featuring the sense-prototype). Such solutions are mostly oriented to highlight the lexical features of the cluster/meaning without considering any element that reflects the cluster's semantics. As a consequence, open challenges are based on the need of comprehensive description techniques capable of capturing both lexical and semantic aspects such as position in text, semantics, or co-occurrences across different documents. \\

\textbf{Word meaning evolution.} 
In shared competitions, the reference evaluation framework for \app\ is based on one/two time periods that are considered for shift detection. The extension of the evaluation framework to consider more time periods is an open challenge. In particular, methods and practices of \app\ approaches need to be tested/extended for detecting both short- and long-term semantic shifts and for promoting the design of incremental techniques able to handle dynamic corpora (i.e., corpora that become progressively available).

In this context, a further challenge is about the capability to trace the change of a meaning over multiple time steps (i.e., meaning evolution). As mentioned in Section~\ref{sec:problem}, alignment   techniques can be used to link similar word meanings in different, consecutive time periods. However, such a solution is not completely satisfactory due to possible limitations (e.g., scalability, robustness of alignment) and further research work is needed to better track the meaning evolution over time (e.g.,~\cite{periti2022what}). \\

\textbf{Model stability.} 
Most of the approaches surveyed in this paper are time-oblivious and face the problem of model stability through fine-tuning. Since this practice can be expensive in terms of time and resources, we argue that further research on the development of time-aware approaches is needed, in that, they do not suffer the model stability problem.\\

\textbf{Model bias.}
The solutions to model bias issues presented in Section~\ref{sec:limitations} are language-dependent and they are mainly exploited in approaches based on monolingual contextualised model. Further research work is needed to test the effectiveness of existing solutions also in approaches based on multilingual contextualised models. In addition, we argue that future work should concern the application of denoising and debiasing techniques to both monolingual and multilingual embedding models (e.g.,~\cite{kaneko2021debiasing}) with the aim to improve \app\ performance by reducing orthographic biases regardless of the language(s) on which the models were trained.\\

\textbf{Further challenges} not strictly related to the issues of Section~\ref{sec:limitations} are the following: \\
\textit{Semantic Shift Interpretation.}
Most of the literature papers do not investigate the nature of the detected shifts, meaning that they do not classify the semantic shifts according to the existing linguistic theory (e.g., amelioration, pejoration, broadening, narrowing, metaphorisation, metonymisation, and metonymy)~\citep{campbell2013historical,hock2019language}.
Further studies on the causes and types of semantic changes are needed. These studies could be crucial to detect ``laws'' of semantic shift that describe the condition under which the meanings of words are prone to change. For example, some laws are hypothesised in~\cite{yang2015computational,dubossarsky2015bottom,hamilton2016diachronic}, but later the validity of some of them has been questioned~\citep{dubossarsky2017outta}. Contextualised embeddings could contribute to test the validity of current laws and to propose new ones. To the best of our knowledge, some steps in this direction are only moved in~\cite{hu2019diachronic} for modeling the word change from an ecological viewpoint. \\

\textit{Computational models of meaning change.} Almost all experiments on \app\ are based on BERT embeddings. Although there are open questions about how to maximise the effectiveness of BERT embeddings in different language setups, the effectiveness of BERT for tracing semantic shifts has been extensively investigated. We believe that \app\ should be extended by considering a wider range of contextualised embedding models. Some work explored the effectiveness of ELMo~\citep{kutuzov2020uio,rodina2020elmo}. However, the performance of ELMo in different contexts and setups should be analysed in more detail. Furthermore, it might be worth investigating smaller versions of BERT, like ALBERT~\citep{lan2020albert} and DistilBERT~\citep{sanh2020distilbert}. Further models can also be considered like seq2seq and generative models, which recently showed interesting results in the field of temporal Word-in-Context problem~\citep{lyu2022mllabs}. \\ %The results have shown the plausibility of using generation model for WiC tasks, meanwhile also indicate there’s still room for further improvement
    
\textit{Multilingual models.} 
In past shared competitions on SSD, monolingual models have generally been preferred to multilingual ones. We believe that a systematic comparison of monolingual vs. multilingual models is required to determine scenarios and conditions where the former type of models provides better performance than the latter type or vice-versa. Multilingual embeddings can also contribute to \app\ since they could enable a language-independent semantic shift assessment, meaning that the gold-scores of different languages can be exploited as a whole for the evaluation of a given approach.\\

\textit{Cross-language shift detection}. 
As introduced in~\cite{martinc2020leveraging}, further investigations are required to address the problem of cross-language shift detection. We argue that solutions to such a kind of problem can be also useful for \app\ since they can detect semantic change of \textit{cognates} and \textit{borrowings} (e.g.,~\citep{fourrier2022caveats}), as well as \textit{contact-induced} semantic shifts (e.g., ~\citep{miletic2021detecting})\footnote{In linguistics, cognates are sets of words in different languages that have been inherited in direct descent from an etymological ancestor in a common parent language. Borrowings (or loanwords) are words adopted by the speakers of one language from a different language. Contact-induced semantic shifts are diachronic changes within a recipient language that are traceable to languages other than the direct ancestor of the recipient language and that have spread and are conventionalised within a community speaking the recipient language.}.\\

\textit{Use cases}. So far, detecting semantic shifts through contextualised embeddings is still a theoretical problem not yet integrated in real application scenarios like historical information retrieval, lexicography, linguistic research, or social-analysis. For this reason, further use cases and experiences must be developed and shared.\\

\textit{Context Shift over different domains}.
The attention gained by diachronic semantic shift detection through the use of word embeddings paved the way for modeling other linguistics issues such as the identification of diatopic lexical variation~\citep{Seifart2019contact}, the detection of semantic shifts of grammatical constructions~\citep{fonteyn2020grammar}, or the comparison of how speakers who disagree on a subject use the same words~\citep{garisolerone2022}. The \app\ approaches can be tested and possibly extended to cope with such a kind of linguistics issues.

\section*{Acknowledgments}
We would like to thank Nina Tahmasebi for her valuable comments and constructive feedback on the manuscript.

\bibliographystyle{unsrtnat}
\bibliography{references}  

%%% Uncomment this line and comment out the ``thebibliography'' section below to use the external .bib file (using bibtex) .






% \end{thebibliography}

\end{document}